%% file: main.tex
\documentclass[10pt,twocolumn,letterpaper]{article}

\usepackage{cvpr}              

\usepackage{url}
\input{math_commands.tex}

\usepackage{epsfig}
\usepackage{graphicx}
\usepackage{amsmath}
\usepackage{amssymb}
\usepackage{caption}
\usepackage{multirow}
\usepackage{url}
\usepackage{xcolor}
\usepackage{xspace}
\usepackage{pifont}

\usepackage{wrapfig,lipsum,booktabs}
\usepackage{colortbl}
\definecolor{mygray}{gray}{0.9}
\definecolor{myorange}{rgb}{0.988, 0.702, 0.506}
\definecolor{myyellow}{rgb}{0.992, 0.902, 0.706}
\newcommand{\thickhline}{\noalign{\hrule height 1.8pt}}

\newcommand{\thinerhline}{\noalign{\hrule height 0.3pt}}
\usepackage[most]{tcolorbox}
\usepackage{makecell}

\newcommand{\methodname}{PSHuman\xspace}
\definecolor{cvprblue}{rgb}{0.21,0.49,0.74}
\usepackage[pagebackref,breaklinks,colorlinks,allcolors=cvprblue]{hyperref}

\def\paperID{10342} 
\def\confName{CVPR}
\def\confYear{2025}
\title{PSHuman: Photorealistic Single-image 3D Human Reconstruction 
using Cross-Scale Multiview Diffusion and Explicit Remeshing}

\author{
         Peng Li$^{1}$, ~~
        Wangguandong Zheng$^{2}$,~
        Yuan Liu$^{1}$\footnotemark[2],~
        Tao Yu$^{3}$,~
        Yangguang Li$^{4}$,
        Xingqun Qi$^{1}$
    \\
    \quad\quad
        Xiaowei Chi$^{1}$,
        Siyu Xia$^{2}$,
        Yan-Pei Cao$^{4}$,
        Wei Xue$^{1}$,
        Wenhan Luo$^{1}$\footnotemark[2],
        Yike Guo$^{1}$
    \\
    {~~~~~~~\normalsize $^{1}$HKUST} \quad
    {\normalsize $^{2}$Southeast University} 
    {~~~~~\normalsize $^{3}$Tsinghua University}~~~
    {\normalsize $^{4}$VAST}
    \\
        \tt \small \centering \textbf{\href{https://penghtyx.github.io/PSHuman}{https://penghtyx.github.io/PSHuman}
    }
}

\begin{document}
\maketitle
\input{sections/0_abstract}    
\input{sections/1_intro}
\input{sections/2_related_works}
\input{sections/3_method}

\input{sections/4_experiments}

\input{sections/5_conclusion}

\noindent\textbf{Acknowledgement.} The research was supported by Theme-based Research Scheme (T45-205/21-N) from Hong Kong RGC, Generative AI Research, Development Centre from InnoHK, NSFC (No.62422110) and Guoqiang Institute of Tsinghua University (No.2021GQG0001).

{
    \small
    \bibliographystyle{ieeenat_fullname}
    \bibliography{main}
}
\input{sections/X_suppl}


\end{document}

%% file: math_commands.tex
\usepackage{amsmath,amsfonts,bm}

\def\eqref#1{equation~\ref{#1}}

\def\1{\bm{1}}

\DeclareMathAlphabet{\mathsfit}{\encodingdefault}{\sfdefault}{m}{sl}
\SetMathAlphabet{\mathsfit}{bold}{\encodingdefault}{\sfdefault}{bx}{n}

\DeclareMathOperator*{\argmin}{arg\,min}

%% file: sections/0_abstract.tex
\begin{abstract}
Photorealistic 3D human modeling is essential for various applications and has seen tremendous progress. However, existing methods for monocular full-body reconstruction, typically relying on front and/or predicted back view, still struggle with satisfactory performance due to the ill-posed nature of the problem and sophisticated self-occlusions. In this paper, we propose \textbf{\methodname}, a novel framework that explicitly reconstructs human meshes utilizing priors from the multiview diffusion model. It is found that directly applying multiview diffusion on single-view human images leads to severe geometric distortions, especially on generated faces.
To address it, we propose a cross-scale diffusion that models the joint probability distribution of global full-body shape and local facial characteristics, enabling identity-preserved novel-view generation without geometric distortion. Moreover, to enhance cross-view body shape consistency of varied human poses, we condition the generative model on parametric models (SMPL-X), which provide body priors and prevent unnatural views inconsistent with human anatomy. Leveraging the generated multiview normal and color images, we present SMPLX-initialized explicit human carving to recover realistic textured human meshes efficiently. Extensive experiments on CAPE and THuman2.1 demonstrate \methodname's superiority in geometry details, texture fidelity, and generalization capability. 


\end{abstract}

%% file: sections/1_intro.tex
\section{Introduction}
Photorealistic 3D reconstruction of clothed humans is a promising and widely investigated research domain with significant applications across several industries, including gaming, movies, fashion, and AR/VR~\cite{ma2021pixel, orts2016holoportation}. Traditional methods, which perform multiview stereo and non-rigid registration using multi-camera setups or incorporate additional depth signals, have achieved accurate modeling. However, reconstruction from an in-the-wild RGB image remains an open problem due to sophisticated body poses and complex clothing topology. 

\begin{figure}
    \centering
    \includegraphics[width=0.475\textwidth]{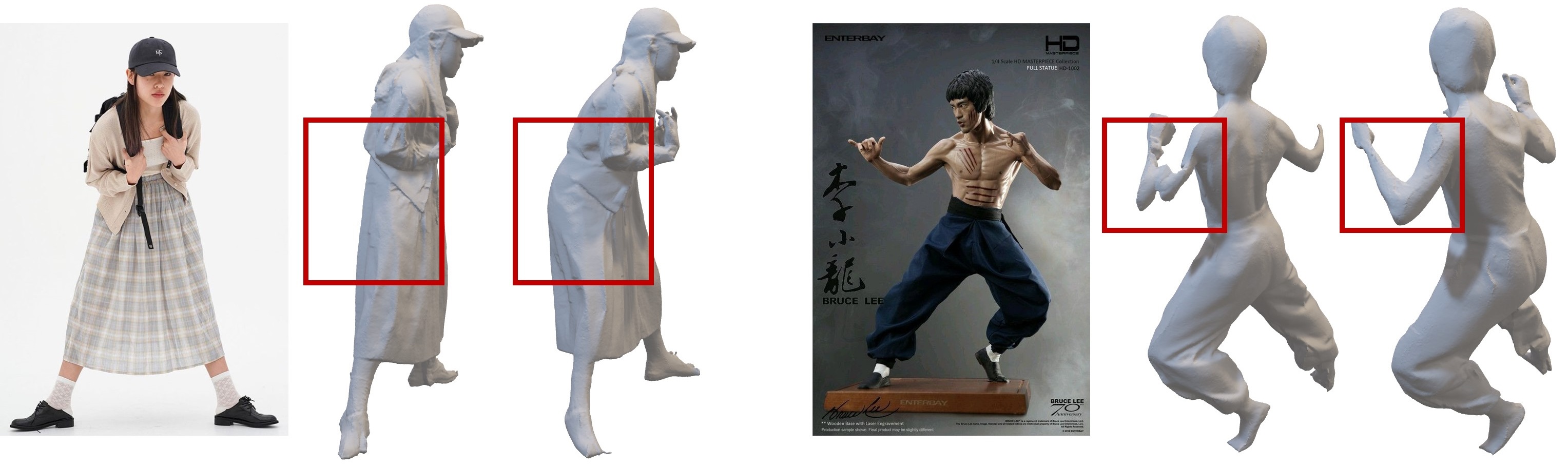}
        \footnotesize\leftline{~~~\quad~~(a)~~\qquad\qquad\qquad\qquad\qquad\qquad\quad~~~(b)}
        \caption{Each triplet contains input (left) and reconstructions of w/o (middle) and w/ (right) SMPL-X condition. Compared with naive diffusion, SMPL-X prior guides handling self-occlusion and improving consistency.}
        \label{fig:abla_smpl}
        \vspace{-10pt}
\end{figure}

Early studies~\cite{saito2019pifu, saito2020pifuhd, xiu2022icon} utilize implicit functions~\cite{occupancy_network, deepsdf} to recover textured human mesh from a single color or normal image. Despite improvements in monocular ambiguity and postural intricacy, this regression paradigm still falls short in detail loss and novel view artifacts. Recent efforts~\cite{zhang2024sifu,ho2024sith} incorporate generative information, such as predicting a back view, to mitigate these issues. On the one hand, the reconstruction pipelines still follow implicit functions, which exhibit limitations  in capturing high-fidelity geometry and texture details. On the other hand, the introduction of the back view fails to provide enough stereo information to mitigate spatial ambiguity. 


In this study, we aim to tackle these existing challenges by introducing a multiview diffusion model and a normal-guided explicit human reconstruction framework. Different from the front and/or back views by existing methods~\cite{xiu2023econ,ho2024sith}, we explore the direct multiple views generation for robust human modeling. As depicted in Fig.~\ref{fig:pipeline}, \methodname takes a full-body human image as input, followed by a designed multiview diffusion model and an SMPLX-initialized mesh carving module, outputting a textured 3D human mesh.

Specifically, we fine-tune a pre-trained text-to-image diffusion model (such as Stable Diffusion~\cite{stable_diffusion}) to generate multiview color and normal maps conditioned on the input reference. Despite impressive generative performance, this base framework faces two major challenges: 1) \textbf{Unnatural body structures}, where diffusion models struggle to generate reasonable novel views of posed human, often resulting in disproportionate body proportions or missing body parts. As shown in Fig.~\ref{fig:abla_smpl}, this issue arises from the severe self-occlusion in the posed human image and lack of body prior for generative models.
To address this, we propose a SMPL-X conditioned diffusion model, which concatenates renderings of estimated SMPL-X with the input image to provide pose guidance for novel-view generation. This approach constrains the diffusion model to generate consistent views that adhere to human anatomy, even when fine-tuning with as few as $3,000$ human scans. 2) \textbf{Face distortion}, where pre-trained diffusion models often produce distorted and unnatural face details, especially for full-body human input. This problem is attributed to the small size of the face in full-body images, which provides limited information for detailed normal prediction after VAE encoding. To accurately recover face geometry, we propose a body-face cross-scale diffusion framework that simultaneously generates multiview full-body images and local face ones. We also employ a simple yet efficient noise blending layer to enhance face details in global image, guaranteeing both cross-scale and cross-view consistency. Consequently, \methodname generates high-quality and detailed novel-view human images and corresponding normal maps. 

To fully leverage the generated multiview images, we present a SMPLX-initialized explicit human carving module for fast and high-fidelity textured human mesh modeling. Unlike implicit functions that use Multilayer Perceptrons (MLPs) to map normal features to an implicit surface, or BiNI~\cite{bini2022cao} that utilizes variational normal integration to recover 2.5D surfaces, we directly reconstruct the 3D mesh supervised by generated multiview normal maps. In practice, we initialize the human model with predicted SMPL-X, and deform and remesh it with differentiable rasterization~\cite{palfinger2022continuous}. 
As shown in Fig.~\ref{fig_teaser}, \methodname can preserve fine-grained details, such as facial features and fabric wrinkles, and generate natural and harmonious novel views. For texturing on the generated meshes, we first fuse multiview color images using differentiable rendering to mitigate generative inconsistencies, then project them onto the reconstructed 3D mesh. 

The entire reconstruction process takes as short as one minute. It is noted that recent SDS-based methods~\cite{huang2024tech, huang2024humannorm} also achieve state-of-the-art performance in geometry details and appearance fidelity. However, they can only handle simple poses and suffer from time-consuming optimization (\textit{e.g.}, TeCH~\cite{huang2024tech} takes approximately six hours). Conversely, \methodname achieves a balance between precision, efficiency, and pose robustness. 

In summary, our key contributions include:
\begin{itemize}
     \item We introduce \methodname, a novel diffusion-based explicit method for detailed and realistic 3D human modeling from a single image. 
     \item We present a body-face cross-scale diffusion and a SMPL-X conditioned multiview diffusion for high-quality full-body human image generation with high-fidelity face details. 
     \item  We design a SMPLX-initialized explicit human carving module to fast recover textured human mesh based on generated multiview cross-domain images, achieving SOTA performance on THuman2.1 and CAPE datasets. 
\end{itemize}

\label{sec:intro}

%% file: sections/2_related_works.tex
\section{Related Works}
\noindent\textbf{Implicit Human Reconstruction.} Implicit functions have gained significant traction in human reconstruction~\cite{chibane2020ifnet, gropp2020implicit, D-IF} due to their flexibility in handling complex topology and diverse clothing styles. Pioneering works such as PIFu~\cite{saito2019pifu} introduce pixel-aligned implicit functions, mapping 2D image features to 3D implicit surface for continuous modeling. Building upon this, subsequent research incorporates parametric models (e.g., SMPL) to enhance anatomical plausibility and robustness in challenging in-the-wild poses~\cite{he2020geopifu,xiu2022icon,zheng2021pamir,gta}  or for animation-ready modeling~\cite{arch, arch++}. Other efforts enhance geometric details and dynamic stability by introducing normal~\cite{saito2020pifuhd}, depth clues ~\cite{function4d, visrecon}, or decoupling albedo~\cite{phorhuman} from natural inputs. However, these methods struggle with unseen areas due to limited observed information. More recent approaches~\cite{zhang2024sifu, ho2024sith} incorporate predicted side-view images to enhance visualization but still face challenges in balancing quality, efficiency, and robustness.


\noindent\textbf{Explicit Human Reconstruction.} Early research focuses on explicit representation for human reconstruction. Voxel-based methods~\cite{bodynet, deephuman} utilize 3D UNet to predict volumetric confidence occupied by the human body, which demands high memory and often results in compromised spatial resolution, hindering the capture of fine details crucial for realistic representation. As a more efficient alternative, visual hulls~\cite{SiCloPe} approximate 3D shapes by incorporating silhouettes and 3D joints. Another strategy involves using depth ~\cite{Moulding_Humans, FACSIMILE, 2k2k} or normal~\cite{Tex2Shape,xiu2023econ} information to explicitly infer the 3D human body, balancing detail preservation with computational efficiency. Among these, ECON utilizes normal integration and shape completion, achieving extreme robustness for challenging poses and loose clothing. The major limitations lie in sub-optimal geometry and supporting appearance. To address this, we propose to simultaneously recover geometry and appearance with differentiable rasterization under the supervision of multiview normal and color maps predicted by the diffusion model.

\noindent\textbf{Diffusion-based Human Reconstruction.}
 Most recently, Score Distillation Sampling (SDS)~\cite{poole2022dreamfusion} based human generation methods~\cite{liao2023tada,huang2024tech} have achieved SOTA performance.
 However, these approaches often require time-consuming optimization. Drawing inspiration from the advancement of multiview diffusion based 3D generation~\cite{liu2023syncdreamer, wonder3d, li2024era3d, voleti2024sv3d, tang2024mvdiffusionpp}, our work reduces the inference time by directly generating multiple human views for human reconstruction. We further augment human generation capabilities through the introduction of a novel SMPL-X-conditioned cross-scale attention framework. Most related to our work, Chupa~\cite{chupa} also reconstructs with multiview normals.  However, it still depends on optimization-based refinement and does not support image condition and texture modeling.

\label{sec:rw}

%% file: sections/3_method.tex
\section{Method}
\textbf{Overview.} Given a single color image, \methodname recovers a textured human mesh by two primary stages: 1) a body-face cross-scale multiview diffusion conditioned on SMPL-X, which generates multiview full-body cross-domain (color and normal) images and local facial ones (Sec.~\ref{subsec:mv_gen}), 2) an SMPLX-initialized explicit human carving module for modeling 3D textured meshes (Sec.~\ref{subsec:recon}). Different from previous works utilizing front and/or back views,  we follow \cite{li2024era3d, wonder3d} to directly generate six views (front, front left, left, back, right, and front right) for explicit reconstruction, which strike the best balance between computational cost and effectiveness. Since we generate normal maps and images, we use $x$ and $z$ as the raw data and latent for both modalities.

\begin{figure*}
    \centering
    \includegraphics[width=\textwidth]{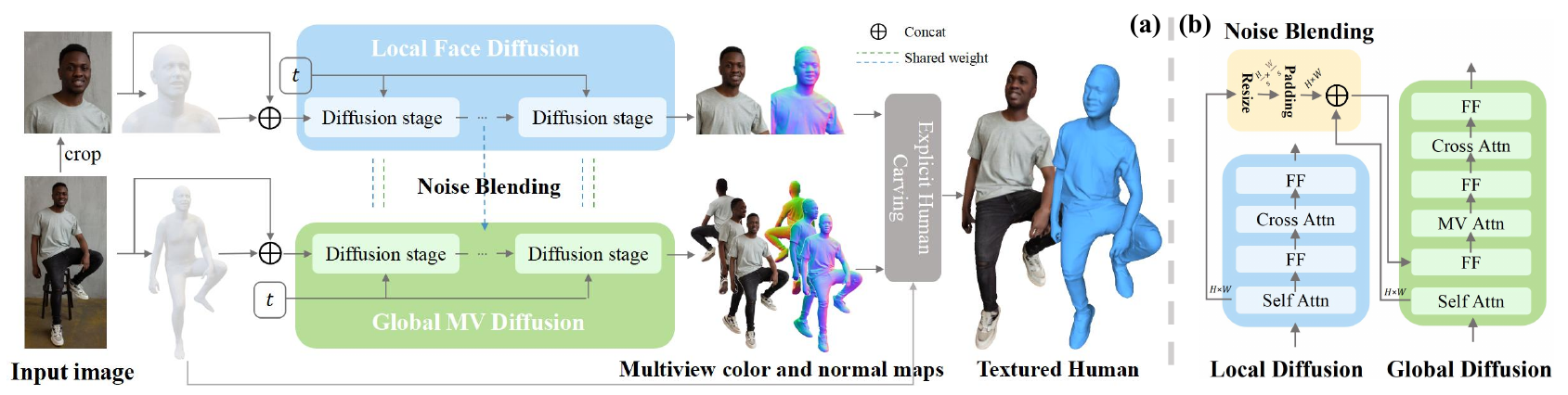}
    \caption{\textbf{(a)} \textbf{Overall pipeline}. Given a single full-body human image, \methodname recovers the texture human mesh by two stages: 1) Body-face enhanced and SMPL-X conditioned multiview generation. The input image and predicted SMPL-X are fed into a multiview image diffusion model to generate six views of global full-body images and front local face images. 2) SMPLX-initialized explicit human carving. Utilizing generated normal and color maps to deform and remesh the SMPL-X with differentiable rasterization. \textbf{(b)} Illustration of joint denoising diffusion block.}
    \label{fig:pipeline}
    \vspace{-10pt}
\end{figure*}

\subsection{Body-face Multiview Diffusion}
\label{subsec:mv_gen}

\subsubsection{Body-face Diffusion} 
\textbf{Motivation.} Simply adopting the multiview diffusion~\cite{li2024era3d,wonder3d} for 3D human reconstruction leads to distorted faces and altered facial identities. Because the face only occupies a small region with a low resolution in the image and cannot be accurately generated by the multiview diffusion model. Since humans are very sensitive to slight changes in faces, such generation inaccuracy of faces leads to obvious distortion and identity changes. This motivates us to separately apply another multiview diffusion model to generate the face at a high resolution with more accuracy.

\noindent\textbf{Forward and reverse processes.} 
We define our data distribution $p(x)$ as the joint distribution of the human face $x^F$ and the human body $x^B$ by
\begin{equation}
    p(\mathbf{x})=p(x^B,x^F)=p(x^B|x^F)p(x^F).
\end{equation}
Then, we follow the DDPM model to define our forward and reverse diffusion process by
\begin{equation}
    q(x_t|x_{t-1})=q(x^B_t|x^B_{t-1},x^F_{t-1})  q(x^F_t|x^F_{t-1}),
\end{equation}
\begin{equation}
    p(x_{t-1}|x_t)=p(x^B_{t-1}|x^B_t, x^F_{t-1})p(x^F_{t-1}|x^F_t),
\end{equation}
where $q$ defines the forward process to add noise to the original data and $p$ defines the reverse process to generate data by denoising. For the forward process, we omit the condition on the $x^F_{t-1}$ and add noises to the face and body images separately by the approximated forward process
\begin{equation}
    q(x_t|x_{t-1})\approx q(x^B_t|x^B_{t-1}) q(x^F_t|x^F_{t-1}).
\end{equation}
Although explicitly defining forward process for $q(x^B_t|x^B_{t-1},x^F_{t-1})$ is feasible for the vanilla diffusion model, it is difficult for the latent diffusion model. We explain this difficulty and the feasibility of this approximation in supplementary material. For the reverse process $p(x_{t-1}|x_t)$, the face diffusion is just a vanilla diffusion model $p(x^F_{t-1}|p^F_{t})$ while the body diffusion model will additionally use the face denoising results as conditions by $p(x^B_{t-1}|p^{B}_t,p^{F}_{t-1})$, as shown in Fig.~\ref{fig:pipeline}(b), which is implemented by the following joint denoising scheme.

\noindent\textbf{Joint denoising.} We utilize a simple but efficient noise blending layer to jointly denoise in body-face diffusion. Specifically, in each self-attention block of UNet, we extract the latent vector of the face branch, resize it with scale $s$, and add it to the face region of the global branch with a weight $w$. Specifically, let us take one of the hidden layers as an example. We denote $h^{B_n}_t$ and $h^F_t$ as hidden vectors of the $n$-th body view and face view at the same attention layer \footnote{Here, we omit the layer subscript for simplicity.} and timestep $t$, the blending operation can be written as 
\begin{align}
h^{B_n}_t &= \begin{cases}
h^{B_1}_t + w \cdot \mathop{RP}(h^F_t, s), & n = 1  \cr
h^{B_n}_t, & n = 2, 3, \dots, N
\end{cases}
\end{align}
where the $\mathop{RP}$ is the resize and padding function, and $w$ is a binary mask of the face region, which is obtained with a face detector or a straightforward cropping strategy. 
The resulting latent vector can be represented by $z^{B_n}_t$ and $z^F_t$. We jointly optimize the body and face distribution with the following loss, 
\begin{align}
    \ell = & ~~~~\mathbb{E}_{t,{z}^F_0,\mathbf{\epsilon}}\left[\|\mathbf{\epsilon} - \mathbf{\epsilon}_\theta (z^F_t, t)\|_2\right] \nonumber \\
    &\ \  + \mathbb{E}_{t,{z}^B_0,{z}^F_0,n,\mathbf{\epsilon}} \left[\|\mathbf{\epsilon}^{(n)} - \mathbf{\epsilon}^{(n)}_\theta (z^B_t,z^F_t,t)\|_2\right],
\end{align}
where $\theta$ is shared weights between face and body views. The noise blending allows the face information to be transferred to novel body views with cross-view attention, improving the overall consistency of generated human images.  

\subsubsection{SMPL-X Guided Multiview Diffusion}
Our multiview diffusion model excels in generating plausible novel views for simple posed images, producing natural human geometry. However, it faces challenges with in-the-wild images that often feature self-occlusions. These occlusions can lead to ``hallucinations" that violate human structural integrity or exhibit inconsistent limb poses. For example, Fig.~\ref{fig:abla_smpl} illustrates two common issues: (a) the model generating upright side views for a bending posture, and (b) inconsistencies in arm regions of side views due to self-occlusion, resulting in failed reconstruction.

To mitigate these impediments, we propose incorporating additional pose guidance into the diffusion process. Our method first estimates the SMPL-X of the input image and renders them from six target viewpoints. We then utilize a pre-trained Variational Autoencoder (VAE) encoder to convert SMPL-X renderings and reference images into latent vectors, which are concatenated with noise samples to serve as input of the denoising UNet. The introduction of these conditional signals constrains the multiview distribution, leading to more accurate and consistent human image generation. This approach significantly enhances the model's generalization capability on complex human poses with self-occlusion.   


\subsection{SMPLX-initialized Explicit Human Carving}
\label{subsec:recon}

\begin{figure}
    \centering
    \includegraphics[width=0.478\textwidth]{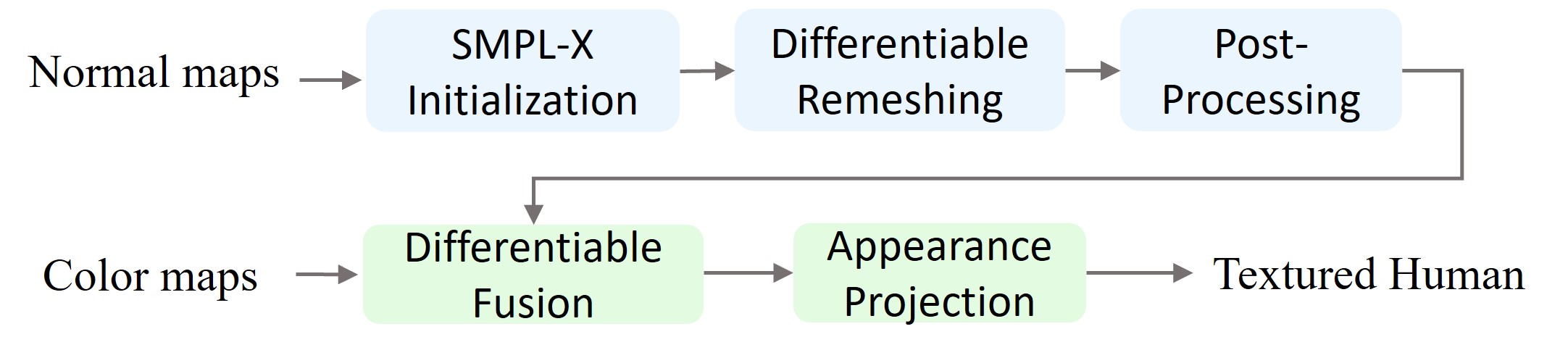}
    \caption{Illustration of our explicit human carving module.}
    \label{fig:reconstruction}
\end{figure}
Following the generation of multiview color and normal images, we elaborate on our SMPLX-initialized human carving module (Fig.~\ref{fig:reconstruction}) to obtain the textured 3D mesh. 

Numerous methodologies have been developed to leverage normal cues for human reconstruction.
However, a significant proportion of them employ implicit functions (\textit{e.g.} MLP) to map the normal feature as implicit surfaces. This process, while effective in certain scenarios, often results in a lack of fine geometric details. Even with BiNI used in ECON, the overall geometry still exhibits a notable degradation. Taking advantage of the multiview consistent normal maps, we opt to fuse it directly with the explicit triangle mesh. Our reconstruction module consists of three main stages: SMPL-X initialization, differentiable remeshing, and appearance fusion. 

\noindent\textbf{SMPL-X initialization.} The process commences with human mesh initialization, utilizing the aforementioned SMPL-X estimation, which provides a strong body prior, effectively mitigating unnecessary face pruning and densification during subsequent geometry optimization. However, it is noteworthy that the generated multiple views may exhibit slight misalignment with the SMPL model due to normalization and recentering procedures. Drawing inspiration from ICON, we optimize SMPL-X's translation $\Tilde{t}$, shape $\Tilde{\beta}$, and pose $\Tilde{\theta}$, parameters to minimize:
\begin{equation}
\label{l_init}
     \Tilde{t}, \Tilde{\beta}, \Tilde{\theta} =  \argmin_{t,\beta,\theta} \sum_{i=1}^N w_i(\|N_i- \hat{N}_i\|_2 + \|S_i-\hat{S}_i\|_2),
\end{equation}
where $w_i$ denotes the confidence of $i$-th view, $\hat{N}_i$ and $\hat{S}_i$ represent the SMPL-X normal and silhouette renderings from predefined views. 

\noindent\textbf{Remeshing with differentiable rasterization.} Given the initial human prior, we utilize differentiable rasterization to carve the details based on observational normal maps. While a common approach involves adding per-vertex displacement to the coarse canonical mesh, this method encounters difficulties when modeling complex details, such as loose clothing. To address this limitation, we directly optimize the SMPL topology, encompassing both vertex positions $V$ and face edges $F$. The optimization procedure iteratively applies vertex displacement and remeshing to the triangle mesh, utilizing the optimizer proposed in~\cite{palfinger2022continuous}. The optimization objective can be written as 

\begin{scriptsize}
\begin{equation}
\label{l_remesh}
    \Tilde{V}, \Tilde{F} =  \argmin_{V, F} \sum_{i=1}^N w_i(\|N_i-\hat{N}_i\|_2 + \|S_i-\hat{S}_i\|_2) + \lambda \sum_j(n_j - n_j^{\text{neig}}),
\end{equation}
\end{scriptsize}\\
where $w_i$ denotes the confidence of $i$-th view, $\hat{N}_i$ and $\hat{S}_i$ represent the normal and silhouette renderings from predefined views, $n_j$ and $n_j^{\text{neig}}$ denote the vertex normal and the average normal of neighboring vertices. The regularization weight $\lambda$ is set to $0.02$. We execute $700$ optimization steps to achieve optimal performance. Following the mesh optimization, we employ Poisson reconstruction~\cite{poisson_recon} to complete minor invisible areas, such as the chin. Additionally, following \cite{xiu2023econ}, we offer the option to replace the hands with the estimated SMPL-X results to enhance visual quality.

\noindent\textbf{Appearance fusion.} Upon obtaining the 3D geometry, our objective is to derive the high-fidelity texture matching the reference image. Despite the availability of multiview images, direct projection onto the mesh results in conspicuous artifacts, arising from the cross-view inconsistency and inaccurate foreground segmentation. To overcome this, we perform texture fusion utilizing the aforementioned differentiable rendering. Specifically, we optimize the per-vertex color $VC$ by minimizing
\begin{equation}
    VC = \arg \min_{vc} \sum_{i=1}^N w_i \| C_i - \hat{C}_i \|_2,
\end{equation}
where $C_i$ and $\hat{C}_i$ represent the rendered and generated color images, respectively. In the majority of cases, this color fusion pipeline suffices to generate high-quality appearances. However, certain
areas may remain unobserved from the predefined six viewpoints. Thus, we finally compute a visibility mask and perform topology-aware interpolation based on KDTree, ensuring comprehensive texture coverage.

%% file: sections/4_experiments.tex
\section{Experiments}

\begin{figure*}
    \centering
    \includegraphics[width=\textwidth]{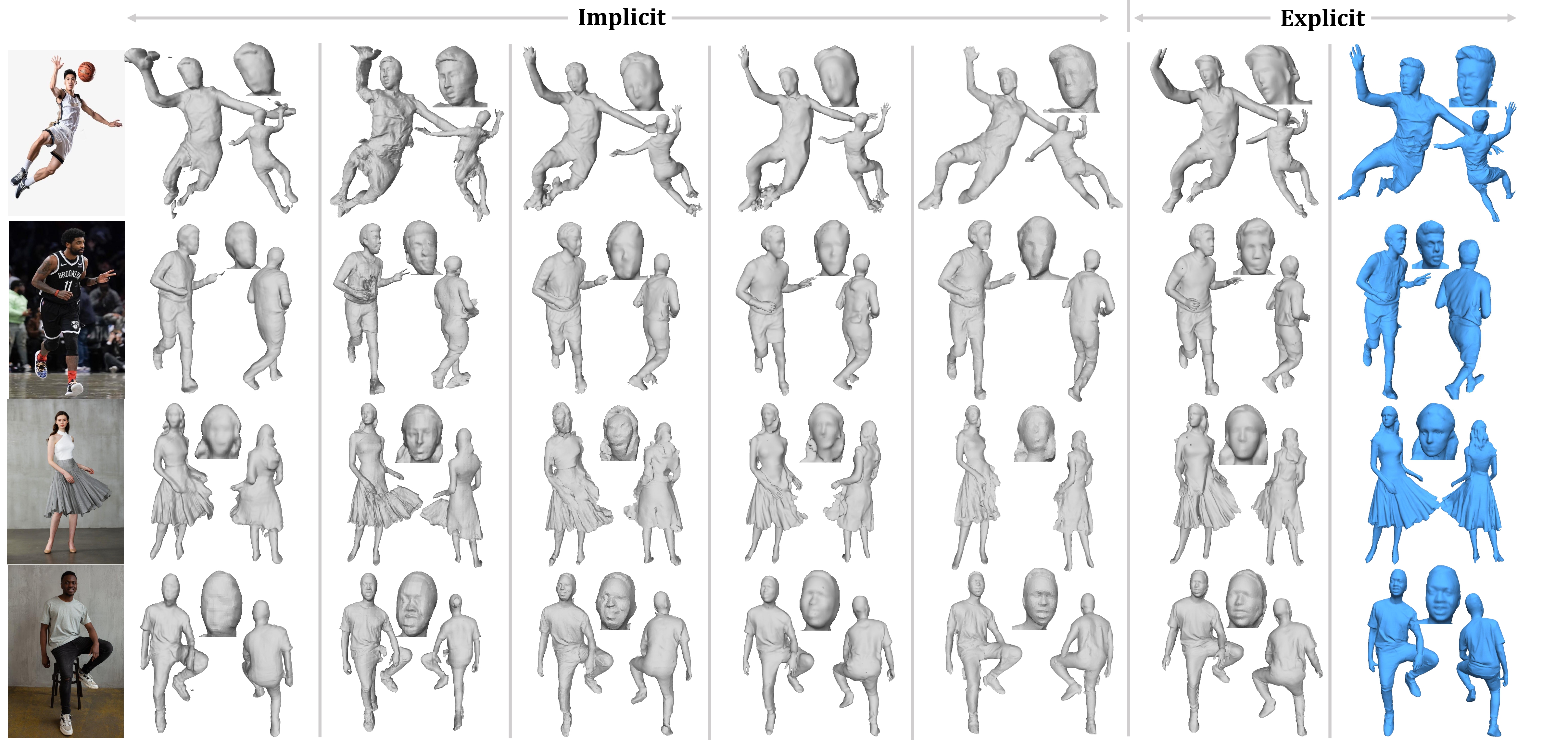}
     \footnotesize \leftline{\quad~~Input~\quad\quad\quad~PaMIR~\quad\quad\qquad~PIFuHD 
    ~~\qquad\quad\quad~~GTA~\qquad\qquad\quad~~~SiFU~\qquad\qquad\qquad~HiLo~\qquad\qquad\quad~~ECON~~\qquad\qquad~~~ Ours}
    \caption{Geometry comparison of PSHuman with \textbf{Implicit} and \textbf{Explicit} methods for 3D human inference from in-the-wild images. Existing methods often struggle with complex poses and loose clothing, leading to issues such as absent body parts, disrupted clothing, and a lack of fine details. In contrast, PSHuman provides a complete shape, detailed facial features, and natural-looking clothing folds. Following \cite{xiu2023econ}, we substitute the hands with SMPL-X models to enhance visual quality.}
    \label{fig:geo_comp}
\end{figure*}

\noindent\textbf{Training and evaluation details.} \methodname builds upon the open-source pre-trained text-to-image generation model, SD2.1-unclip~\cite{stable_diffusion}. Our training is conducted on a cluster of 16 NVIDIA H800 GPUs, with a batch size of 64 for a total of $30,000$ iterations. We adopt an adaptive learning rate schedule, initializing the learning rate at 1e-4 and decreasing it to 5e-5 after $2,000$ steps. The entire training process spans approximately three days. 
Regarding the reconstruction module, we perform SMPL-X alignment, geometry optimization, and texture fusion for $700$, $100$ and $100$ steps, respectively, with corresponding learning rates of $0.3$, $0.001$, and $0.0005$. 
For appearance evaluation~\cite{zhang2024sifu}, we render color images from four viewpoints at azimuths of \{$0^\circ, 90^\circ, 180^\circ, 270^\circ$\} relative to the input view. 

\noindent\textbf{Dataset.} We conduct experiments on widely used 3D human datasets, including THuman2.1~\cite{function4d}, CustomHumans~\cite{CustomHumans} and CAPE~\cite{Ma_cape}. Specifically, our training dataset comprises $2,385$ scans from THuman2.1 and $647$ scans from CustomHumans. These datasets are selected due to their provision of SMPL-X parameters. For quantitative evaluation, we utilize the remaining $60$ scans ($0447$-$0486$, $0492$-$0511$) from THuman2.1 and $150$ scans from CAPE. Following ICON’s partitioning criteria, we subdivide CAPE into "CAPE-FP" (50 samples) and "CAPE-NFP" (100 samples) to assess generalization in real-world scenarios.

\noindent\textbf{Metric.} To assess reconstruction capability, we employ three primary metrics: 1-directional point-to-surface (\textbf{P2S}), $L_1$ Chamfer Distance (\textbf{CD}), and Normal Consistency (\textbf{NC}).
For appearance evaluation, we utilize peak signal-to-noise ratio (\textbf{PSNR})~\cite{wang2004image}, structural similarity index (\textbf{SSIM})~\cite{zhang2018unreasonable}, and learned perceptual image patch similarity (\textbf{LPIPS})~\cite{zhangUnreasonableEffectivenessDeep2018}.

\begin{table*}[bt]
\caption{Quantitative comparison of geometry quality. For the setting of `w/o SMPL-X body prior', we utilize PIXIE to estimate SMPL parameters for other baseline methods while omitting SMPL estimation for our approach. Specifically, we retrain the diffusion model by removing the SMPL-X conditioning and initialize human mesh with a unit sphere during mesh carving. For `w/ SMPL-X body prior', ground-truth SMPL-X models are used to avoid the impact of pose estimation errors on the evaluation. The units for Chamfer and P2S are in cm. The top two results are colored as \colorbox{myorange}{first} \colorbox{myyellow}{second}.}
\vspace{-10pt}
\label{table:geometry_comparision}
\begin{center}
\large
\resizebox{\textwidth}{!}{%
\setlength\tabcolsep{10pt} 
\renewcommand{\arraystretch}{1.4} 
\begin{tabular}{cc|ccc|ccc|ccc}

\rowcolor{mygray}
\thickhline
{} & {} & \multicolumn{3}{c|}{CAPE-NFP} & \multicolumn{3}{c|}{CAPE-FP} & \multicolumn{3}{c}{THuman2.1} \\
\rowcolor{mygray}
Method & Venue & Cham. Dist $\downarrow$ & P2S $\downarrow$ & NC $\uparrow$ & Cham. Dist $\downarrow$ & P2S $\downarrow$ & NC $\uparrow$ & Cham. Dist $\downarrow$ & P2S $\downarrow$ & NC $\uparrow$ \\
\thickhline
\multicolumn{10}{c}{w/o SMPL-X body prior} \\
\hline
PIFu    & ICCV 2019       & 3.2524  & 2.5469	& 0.7624		 & 1.8367  & 1.7582  & 0.8573		 & 1.2071  & 1.1299  & 0.7681  \\
PIFuHD  & CVPR 2020       & 2.9749  & 2.3677    & 0.7658 		 & 1.5211  & 1.4834  & 0.8712 		 & 0.9935  & 0.9647  & 0.7890 \\
PaMIR   & TPAMI 2021      & 7.1577  & 3.3832    & 0.6345		 & 6.0114  & 3.2877  & 0.6737		 & 1.0875  & 1.0144  & 0.7939 \\
ICON   & CVPR 2022        & 2.6983  & 2.3911 	& 0.7958   		 & 2.1331  & 2.0359  & 0.8364  		 & 1.1199  & 1.0925  & 0.7810 \\
ECON   & CVPR 2023        & 3.1086  & 2.6044  	& 0.7722	     & 2.5394  & 2.4336  & 0.8128		 & 1.2500  & 1.1469  & 0.7643  \\
GTA    & NeurIPS 2023     & 2.7387  & 2.4722 	& 0.7875		 & 2.2543  & 2.1889  & 0.8247		 & 1.0612  & 1.0389  & 0.7857 \\
SIFU   & CVPR 2024        & 2.7884  & 2.4792	& 0.7877  		 & 2.1695  & 2.1107  & 0.8310		 & 1.0774  & 1.0586  & 0.7871 \\
HiLo   & CVPR 2024    & 2.6507   & 2.3037 	& 0.7987		 & 2.2735  & 2.1345  & 0.8308 		 & 1.1241   & 1.0519  & 0.7784 \\
SITH    & CVPR 2024       & 2.8735  & 2.1226    & 0.7804 		 & 2.1140  & 1.6754  & 0.8337	     & 0.9661  & 0.9034  & 0.7832 \\
\thinerhline
\textbf{Ours}   & -       & 2.1625  & 1.6675	 & 0.8226 		 & 1.3615  & 1.1308  & 0.8844  		 & 0.6609  & 0.5993  & 0.8310 \\

\hline
\multicolumn{10}{c}{w/ SMPL-X body prior} \\
\hline

ICON   & CVPR 2022       & \cellcolor{myyellow} 1.5511   & \cellcolor{myyellow} 1.1967 	& \cellcolor{myyellow} 0.8572  		 & \cellcolor{myyellow} 0.9951  & \cellcolor{myyellow} 0.8864  & \cellcolor{myorange} 0.9190 		 & 0.6146  & 0.5934  & 0.8493 \\
ECON   & CVPR 2023       & 1.8524   & 1.5706 	& 0.8392	     & 1.1761  & 1.1352  & 0.8969 		 & 0.6725  & 0.6331  & 0.8362 \\
GTA    & NeurIPS 2023    & 1.8853   & 1.4902	& 0.8260		 & 1.1484  & 0.9914  & 0.9011 		 & 0.5791  & 0.5587  & 0.8491 \\
SIFU   & CVPR 2024       & 1.5742   & 1.2777	& 0.8529  		 & 1.0535  & 0.9674  & 0.9024		 & \cellcolor{myyellow} 0.5754  & \cellcolor{myyellow} 0.5576  & \cellcolor{myyellow} 0.8500 \\
HiLo   & CVPR 2024    & 1.5613   & 1.2146	& 0.8547		 & 1.1246  & 0.9847  & 0.9031 		 & 0.5977  & 0.5892  & 0.8405 \\
SITH   & CVPR 2024    & 1.8118   & 1.5201	& 0.8345		 & 1.1839  & 1.1573  & 0.8870 		 & 0.6474  & 0.5810  & 0.8264 \\

\thinerhline
\textbf{Ours}   & -      
& \cellcolor{myorange} \textbf{0.9688}   & \cellcolor{myorange} \textbf{0.8675}	 & \cellcolor{myorange} \textbf{0.8799}		 
& \cellcolor{myorange} \textbf{0.7811}   & \cellcolor{myorange} \textbf{0.6984}  & \cellcolor{myyellow} \textbf{0.9136} 		 
& \cellcolor{myorange} \textbf{0.4399}   & \cellcolor{myorange} \textbf{0.4077}  & \cellcolor{myorange} \textbf{0.8504} \\
\thickhline
\end{tabular}
}
\end{center}
\large
\end{table*}

\subsection{Comparisons}
\textbf{Baselines.} We conducted a comprehensive comparison of our method against state-of-the-art single-view human reconstruction approaches, including PIFu~\cite{saito2019pifu}, PIFuHD~\cite{saito2020pifuhd}, PaMIR~\cite{zheng2021pamir}, ICON~\cite{xiu2022icon}, ECON~\cite{xiu2023econ}, GTA~\cite{gta}, SiFU~\cite{zhang2024sifu}, HiLo~\cite{yang2024hilo}, and SiTH~\cite{ho2024sith}. For SMPL-based methods, we utilize PIXIE~\cite{yu2021pixelnerf} for estimation.  We also report the results with ground-truth SMPL-X to isolate the impact of pose estimation errors.

\noindent\textbf{Comparison of geometry quality.} Leveraging consistent multiview images, our method exhibits superior geometric quality compared to existing approaches, particularly in scenarios without SMPL-X body prior (Tab.~\ref{table:geometry_comparision}). Unlike other template-based methods, which are susceptible to SMPL-X prediction errors, our method supports template-free training, thereby offering enhanced generalization capability. When incorporating the body prior, our method consistently outperforms previous works, demonstrating unprecedented accuracy on complex posed humans. The qualitative comparison in Fig.~\ref{fig:geo_comp} also showcases the superiority of \methodname, featuring with complete shape, detailed face and natural-looking clothing folds.

\begin{table}
\centering
\caption{Quantitative comparisons of face reconstruction.}
\resizebox{0.95\linewidth}{!}{%
\begin{tabular}{c|cccccc}
\thickhline
\rowcolor{mygray}
Method &  Cham. Dist $\downarrow$ & P2S $\downarrow$ & NC $\uparrow$  &  PSNR $\uparrow$ & SSIM $\uparrow$ & LPIPS $\downarrow$ \\
\thickhline
ECON & 0.624 & 0.570 & 0.837 & - & - & - \\
SIFU & 0.535 & 0.527 & 0.853 & 18.86 & 0.790 & 0.093 \\
SITH & 0.610 & 0.563 & 0.858 & 17.93 & 0.827 & 0.110 \\
\hline
w/o local & 0.524 & 0.503 & 0.867 & 19.67 & 0.832 & 0.093 \\
w/o noise blender & \cellcolor{myyellow} 0.447 & \cellcolor{myyellow} 0.422 & \cellcolor{myyellow} 0.904 & \cellcolor{myyellow} 20.85 & \cellcolor{myyellow} 0.877 & \cellcolor{myyellow} 0.075 \\
\textbf{Ours} & \cellcolor{myorange} 0.423 & \cellcolor{myorange} 0.397 & \cellcolor{myorange} 0.924 & \cellcolor{myorange} 20.97 & \cellcolor{myorange} 0.896 & \cellcolor{myorange} 0.071 \\
\thickhline
\end{tabular}}
\label{tab:face_eval}
\end{table}

\begin{table}[htbp]
\centering
\begin{minipage}{0.45\linewidth} 
\caption{Quantitative comparison of appearance rendering on THuman2.1 subset.}
\setlength\tabcolsep{4pt}
\small 
\label{table:appearance_comparision}
\resizebox{\linewidth}{!}{
\begin{tabular}{c|ccc}
\rowcolor{mygray}
\thickhline
Method  & PSNR $\uparrow$ & SSIM $\uparrow$ & LPIPS $\downarrow$  \\
\thickhline
PIFu             & 19.3957  & 0.8327  & 0.1001 \\
PaMIR          	 & 19.4130  & 0.8324  & 0.0988 \\
GTA      		 & \cellcolor{myyellow} 19.6071  & \cellcolor{myyellow} 0.8338  & 0.0989 \\
SIFU             & 19.4417  & 0.8307  & \cellcolor{myyellow} 0.0985 \\
SITH             & 18.4580  & 0.8200  & 0.1004 \\
\hline
\textbf{Ours}    & \cellcolor{myorange} \textbf{20.8548}   & \cellcolor{myorange} \textbf{0.8636}	 & \cellcolor{myorange} \textbf{0.0764}		 \\
\thickhline
\end{tabular}
}
\end{minipage}%
\hspace{0pt} 
\begin{minipage}{0.48\linewidth}
\caption{Evaluation of robustness to SMPL-X estimation on THuman2.1 subset.}
\label{tab:robustness_smplx}
\small
\centering
\setlength\tabcolsep{4pt} 
\resizebox{\linewidth}{!}{ 
\begin{tabular}{c|ccc}
\rowcolor{mygray}
\thickhline
Method  & Cham. Dist $\downarrow$ & P2S $\downarrow$ & NC $\uparrow$  \\
\thickhline
ICON             & 0.7827  & 0.6463  & \cellcolor{myyellow} 0.8401 \\
ECON             &  0.8022 & 0.6742 & 0.8327 \\
GTA             & 0.6631  & 0.6473 & 0.8368\\
SIFU             & 0.6672 & 0.6488 & 0.8302 \\
SITH             & \cellcolor{myyellow} 0.6427 & \cellcolor{myyellow} 0.6393 & 0.8241 \\
\hline
\textbf{Ours}    & \cellcolor{myorange} \textbf{0.5574}   & \cellcolor{myorange} \textbf{0.5377}	 & \cellcolor{myorange} \textbf{0.8417}		 \\
\thickhline
\end{tabular}
}
\end{minipage}
\end{table}

\begin{figure}
    \centering\includegraphics[width=\linewidth]{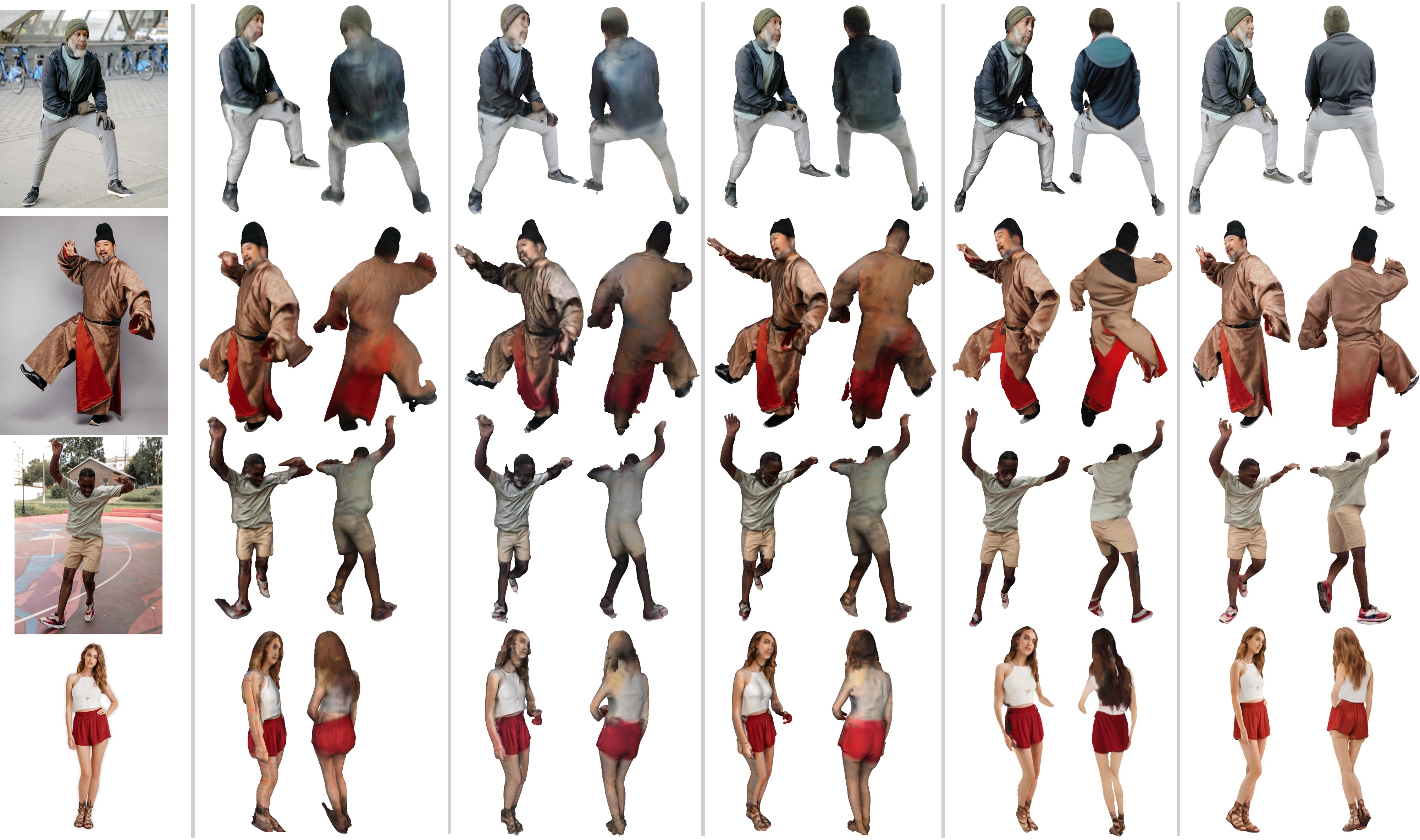}
      \footnotesize \leftline{\quad~Input~\quad~PaMIR 
~~\qquad~GTA~\qquad~~~~~SIFU~\footnote{We only compare with the coarse texture because the refinement code is unavailable. }~\quad~~~~SiTH~\qquad~ Ours}
    \caption{Appearance comparisons with methods which produce texture. Our method could reconstruct realistic and reasonable appearance of side and back views.}
    \label{fig:tex_comp}
    \vspace{-20pt}
\end{figure}

\noindent\textbf{Comparison of appearance quality.}
Quantitative evaluations in Tab.~\ref{table:appearance_comparision} reveal that \methodname outperforms existing methods across multiple metrics, achieving the highest PSNR, SSIM as well as the lowest LPIPS. Qualitatively, as illustrated in Fig.~\ref{fig:tex_comp}, \methodname produces highly consistent appearances on novel viewpoints, including natural and realistic reconstruction for posterior regions. In contrast, existing methods exhibit various limitations such as blurred colors and inconsistent artifacts in unseen views.

\noindent\textbf{Comparisons of face quality.} To highlight the effectiveness of our introduced cross-scale diffusion for face reconstruction, we use the head vertices of SMPL-X to crop the reconstructed head following ECON.  Specifically, we first construct a KD-tree based on SMPL-X to query the generated mesh, subsequently filtering out the vertices adjacent to the head of SMPL-X. Tab.~\ref{tab:face_eval} presents the quantitative comparisons with SOTA methods. 

\subsection{Ablation Study}
\textbf{Effectiveness of SMPL-X condition.} In Fig.~\ref{fig:abla_smpl}, we show the geometry reconstructed by the models trained without SMPL-X condition and with SMPL-X condition. In Fig.~\ref{fig:abla_smpl}(a), it is observed that the naive diffusion model struggles to `imagine' the pose of a bending human image. Conversely, the SMPL-X provides a strong pose prior to guide the model to generate reasonable side views, leading to better reconstruction. In Fig.~\ref{fig:abla_smpl}(b), the diffusion model fails to generate consistent multiple views due to self-occlusion, resulting in artifacts near the arm regions. The SMPL-X guidance effectively enhances consistency, facilitating the complete human body.

\begin{figure}
    \centering
    \includegraphics[width=0.9\linewidth]{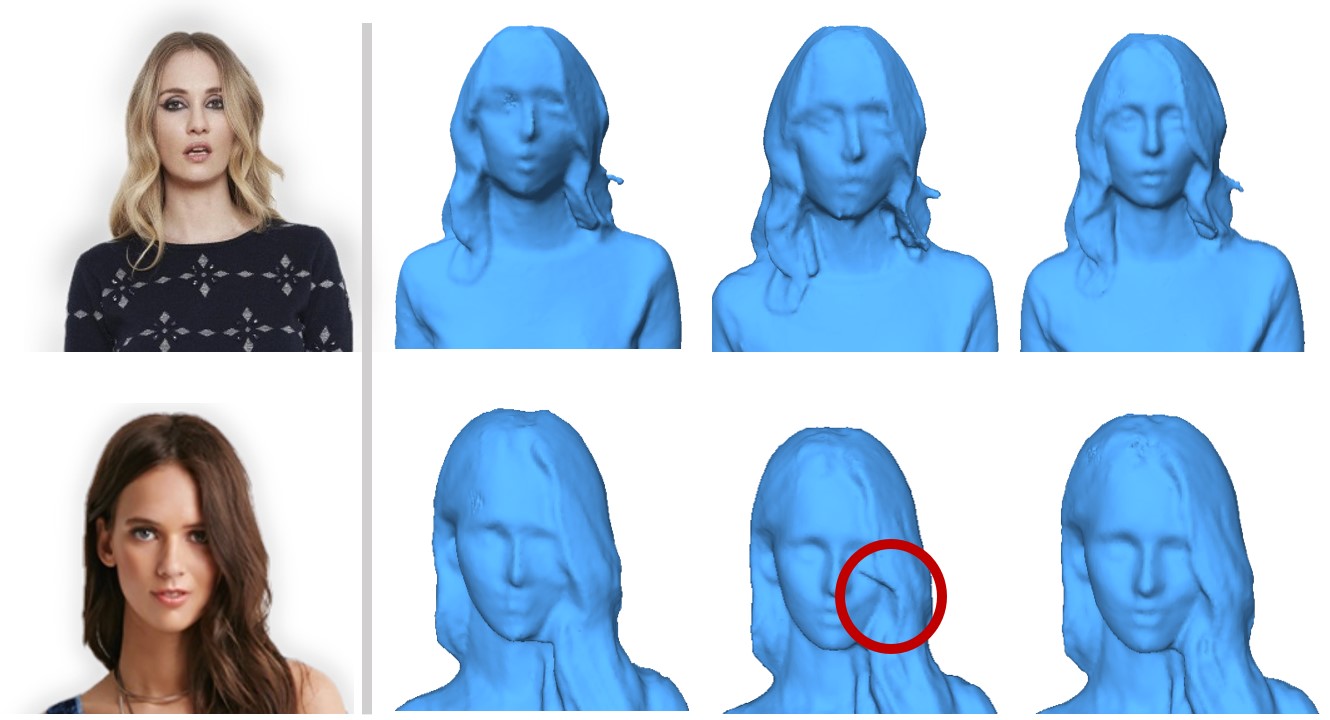}
    \footnotesize\leftline{~~\qquad\quad~Input~\qquad\qquad~w/o local~~~w/o noise blendering~\quad~Ours}
        \caption{Ablation study of the cross-scale diffusion (CSD). The CSD allows sharp face recovery and keeps the identity consistent with the reference input.} 
    \label{fig:abla_lem}
    \vspace{-10pt}
\end{figure}

\noindent\textbf{Effectiveness of cross-scale diffusion (CSD).} In Tab. \ref{tab:face_eval}, we provide the results by removing the local face branch (\textbf{w/o local}) and noise blending (\textbf{w/o noise blending}), respectively. Our method, incorporating both components, achieves the highest performance, 
as shown in Fig.~\ref{fig:abla_lem}. Notably, the setting without noise blending also generates the local face image. However, the reconstructions exhibit minor artifacts or over-smoothness. We attribute it to the inconsistency among global and local images. In contrast, the noise blending allows the information exchange, thereby enhancing overall consistency.  

\begin{figure}
    \centering
    \includegraphics[width=0.95\linewidth]{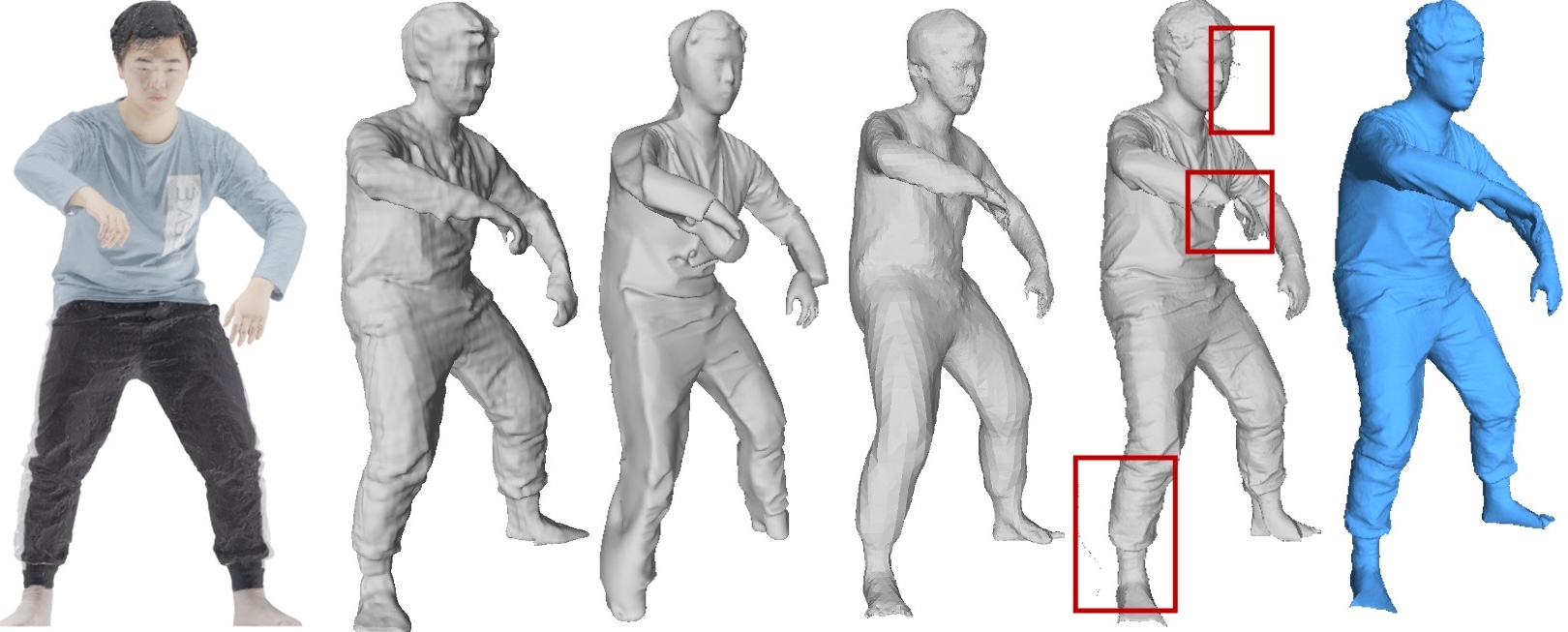}
    \footnotesize\leftline{~~\quad~Input~\qquad\quad~~NeuS~\quad~~~BiNI~~~~~SMPLX-D~~Naive Remesh~~Ours}
    
    \caption{Ablation of our human carving module.}
    \label{fig:abla_recon}
    \vspace{-10pt}
\end{figure}

\begin{figure}
    \centering
    \includegraphics[width=\linewidth]{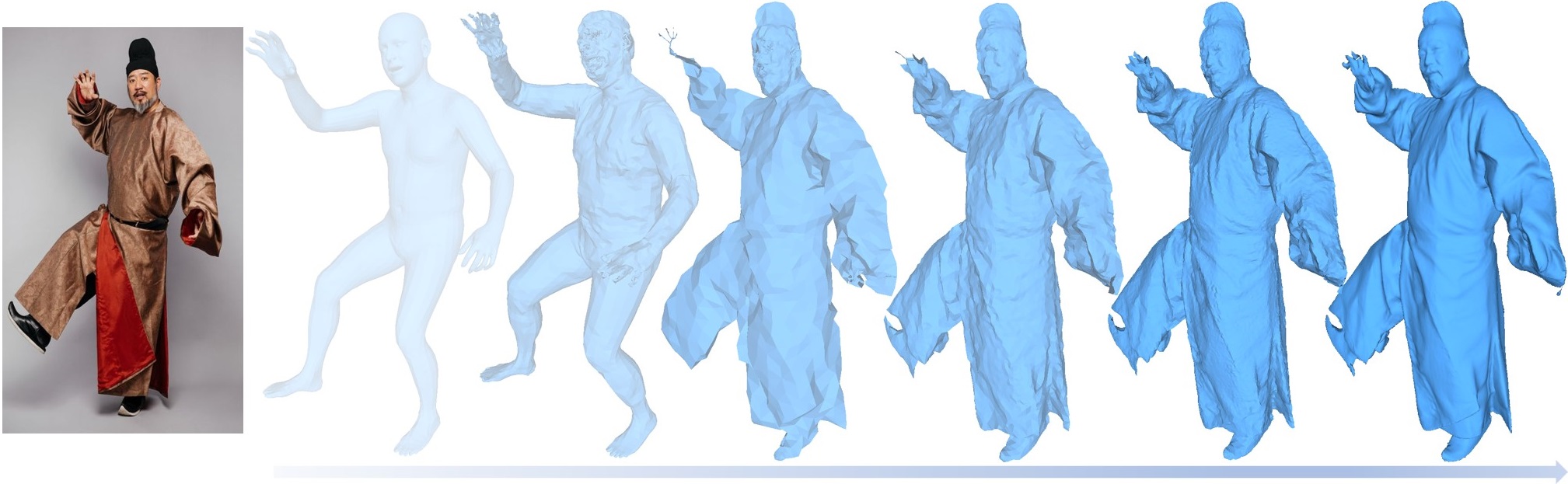}
    \caption{Visualization of mesh carving of a posed human image.}
    \label{fig:carving_process}
\end{figure}


\noindent\textbf{Effectiveness of mesh carving module.} We assess the efficacy of our reconstruction module by substituting the remeshing step with alternative methods, specifically NeuS and BiNI. As illustrated in Fig.~\ref{fig:abla_recon}, the resulting geometries exhibit notable deficiencies or failures to capture fine geometric details. Note that we employ the normal maps, generated by our diffusion model, across all methods to mitigate potential errors arising from normal prediction discrepancies. Moreover, ``naive remeshing" refers to remeshing with SMPL-X initialization but without multiview-guided SMPL-X alignment, resulting in subtle artifacts caused by misalignment between the initial SMPL-X mesh and the multi-view observations. Our reconstruction module effectively addresses these issues. Finally, we show an example across remeshing process for better understanding in Fig.~\ref{fig:carving_process}. 

\noindent\textbf{Robustness to SMPL-X estimation.} We assess the robustness of template-based approaches to SMPL-X estimation errors in Tab.~\ref{tab:robustness_smplx}. Following SIFU, we introduce random noise with a variance of $0.05$ to both the pose and shape parameters of the ground-truth SMPL-X model. The results demonstrate the robust reconstruction capabilities of our approach. Furthermore, the efficacy of our method in real-world scenarios is evidenced by the additional results presented in supplementary materials.

\noindent\textbf{Comprehensive quantitative ablation.} We further conducted comprehensive ablation studies on a subset of $20$ samples from the ``CAPE-NFP" dataset. Tab.~\ref{tab:ablations} quantitatively illustrates the impact on Chamfer Distance when individual components are removed or replaced. It is observed that the SMPL-X condition contributes significantly to reconstruction accuracy. While CSD yields a modest reduction in geometric error, it substantially improves visualization quality and identity fidelity, as evidenced in Fig. \ref{fig:abla_lem}. Furthermore, our reconstruction method, which employs SMPLX-guided differentiable remeshing, demonstrates superior reconstruction performance compared to the BiNI and inpainting pipeline utilized in ECON. 

\begin{table}
\centering
\caption{Comprehensive ablation study of core designs w.r.t full body reconstruction performance.}
\resizebox{\columnwidth}{!}{%
\begin{tabular}{cc|ccc|c}
\thickhline
\multicolumn{2}{c|}{Diffusion} & \multicolumn{3}{c|}{Reconstruction} & \multirow{2}{*}{CD$\downarrow$} \\
CSD & SMPLX-Cond. & Remeshing & SMPLX-ECON & SMPLX-Remeshing  \\
\thickhline
\ding{56} & \ding{56} & \ding{52} & \ding{56} & \ding{56}   & 1.4920  \\
\ding{52} & \ding{56} & \ding{52} & \ding{56} & \ding{56}   & 1.4370  \\
\ding{52} & \ding{52} & \ding{52} & \ding{56} & \ding{56}   & 1.0938  \\
\ding{52} & \ding{52} & \ding{56} & \ding{52} & \ding{56}   & 1.2630  \\
\ding{52} & \ding{52} & \ding{56} & \ding{56} & \ding{52}   & \textbf{0.9597} (Ours)   \\
\thickhline
\end{tabular}%
}
\label{tab:ablations}
\end{table}

%% file: sections/5_conclusion.tex
\section{Conclusion}

\begin{figure}
    \includegraphics[width=\linewidth]{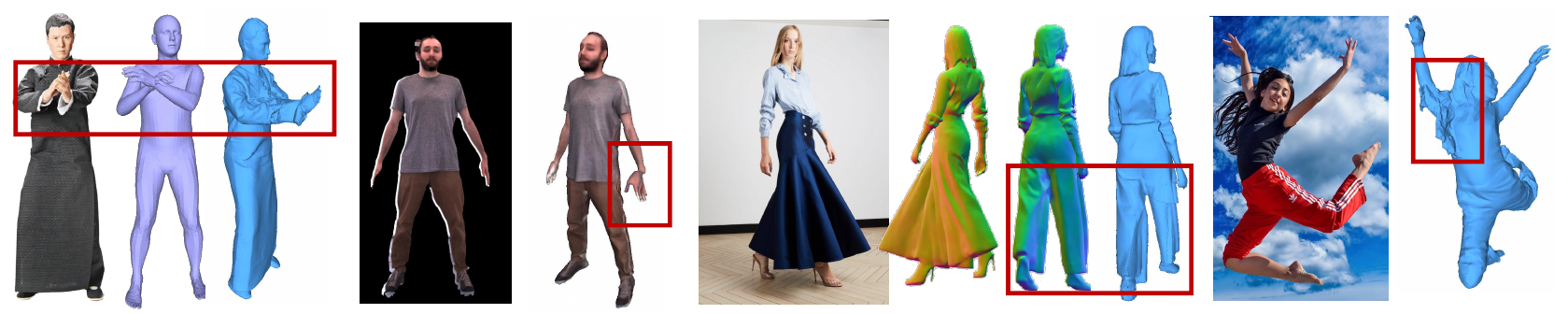}
    \scriptsize\leftline{A.~Inaccurate pose B.~Stitching artifacts~C.~Wrong generation ~\quad~D.~Loose hair
}
    \caption{Failure cases of \methodname. }
    \label{fig:failure_case}
    \vspace{-10pt}
\end{figure}

\noindent\textbf{Limitations.} 
Although \methodname achieves high-quality single-view human reconstruction, it shares certain limitations with previous template-based approaches as shown in Fig.~\ref{fig:failure_case}. First, pose estimation errors (A, B) have a cascading effect on subsequent generation and reconstruction, impacting overall accuracy. In addition, wrong novel-view generation (C) may result in unreasonable geometry. Finally, diffusion models struggle to generate consistent subtle details, such as loose hair and hands (D), which results in suboptimal reconstruction.

\noindent\textbf{Conclusion.} We present PSHuman, a novel framework that significantly improves geometric and appearance quality in single-image human reconstruction. We investigate direct multiview human generation conditioned on SMPL-X, enabling explicit and robust human reconstruction. Our body-face cross-scale diffusion model enhances the modeling of high-fidelity 3D human faces, while our multiview-guided explicit carving module ensures intricate details from generated images. Experiments demonstrate that PSHuman's superiority against existing methods.

%% file: sections/X_suppl.tex
\clearpage
\setcounter{page}{1}
\maketitlesupplementary

\section{Discussions about face-body cross-scale diffusion}
\label{sec:app-diffusion}
\textbf{Difficulty in implementing dependent forward process}. In the dependent forward process $q(x^B_t|x^B_{t-1},x^F_{t-1})$, we know that the face region of $x^B$ corresponds to $x^F$. Since we have defined $p(x^F_t|x_{t-1}^F)$ by adding noises to $x^F_{t-1}$, it is natural to get $x^B_t$ by replacing the pixel values in the face region of $x^B_t$ with $x^F_t$ and just adding noises to the remaining image regions of $x^B_{t-1}$. However, since we adopt a latent diffusion model (Stable Diffusion)~\citet{rombach2022high} here, the pixels of tensors in the latent spaces are not independent of each other so the replacing operation is not valid here. This brings difficulty in separating the face regions in the latent space to explicitly implement the dependent forward process for adding noises.

\noindent\textbf{Rationale of approximated forward process}. Our rationale for adding noises to the face and the body separately is that the process is similar to multiview diffusion. We can regard the face image and the body image as just two images captured by cameras with different camera positions and focal lengths. In this case, the body-face cross-scale diffusion is a special case of multiview diffusion. In a multiview diffusion, we add noises to multiview images separately so that we can also add noises to the body image and face image separately but consider the dependence in the reverse process.

\section{Implementation Details}
\label{sec:supp_details}

\textbf{Preprocessing.} Our training datasets include scans from THuman2.1 and CustomHumans. For each human model, and the corresponding SMPL-X model, we
render 8 color and normal images with alpha channel around the yaw axis, with a $45^\circ$ interval and a resolution of $768\times768$. Due to the random face-forward direction, we employ insightface~\citet{deng2018menpo} for face detection, utilizing only viewpoints containing clear facial characteristics for training. 

\noindent\textbf{Choice of generated views.} As mentioned in the main paper, \methodname generates 6 color and normal images from front, front-right, right, back, left, and front-left views for the trade-off between effectiveness and training workload. To guarantee the generation alignment, we horizontally flip the left and back views during training.  In Fig.~\ref{fig:abla_view}, we present the results reconstructed using only two-view (front and back) or four-view (front, right, back, left) normal maps. Since there is a lack of depth in information, optimizing geometry with fewer views leads to severe artifacts, such as incomplete or unnatural human structures. In contrast, it is evident that the artifacts are reduced when using six views.

\noindent\textbf{Diffusion block.} As illustrated in Fig.3(b) of the main paper, our diffusion block comprises two branches. The local diffusion inherits from stable diffusion (SD2.1-Unclip)~\cite{stable_diffusion}, including self attention, cross attention and feed-forward layers, while the global attention contains an additional multi-view attention layer introduced in Era3D~\cite{li2024era3d}. The global attention is conditioned on the local branch via the noise blending layer. We feed the embeddings of text prompt "a rendering image of 3D human, [V] view, [M] map." into the denoising blocks via cross attention, where [V] is chosen from {"front", "front right", "right", "back", "left", "front left", "face"} and [M] represents "normal" or "color".

\noindent\textbf{Inference details.} Given a human image, we first remove the background with rembg~\cite{rembg} and then resize the foreground to 720$\times$720. Finally, we pad it to 768$\times$768 and set the background to white. Due to the alignment between of processed input image and the generated front color image, we use the former and other generated images in the following reconstruction.  

\section{More experiments}
\label{subsec:A3}
\textbf{Inference time.}
In Tab.~\ref{tab:time}, we report the detailed inference time of the whole pipeline, including preprocessing (SMPL-X estimation and SMPL-X image rendering), diffusion, geometry reconstruction (SMPL-X initialization and remeshing) and appearance fusion. 

\begin{table}
    \centering
    \caption{Inference time of the reconstruction module.}
    \resizebox{\columnwidth}{!}{%
    \begin{tabular}{ccccc}
    \toprule
        Pipeline & Pre-processing & Diffusion & Geo. Recon. & Appearance Fusion \\
     \midrule
        Time~/~s & 7.2 & 17.6 & 23.3 & 6.0 \\ 
    \bottomrule
    \end{tabular}}
    \label{tab:time}
\end{table}

\begin{table}
    \caption{User study w.r.t reconstruction quality and novel-view consistency.}
    \label{tab:user_study}
    \centering
    \resizebox{\columnwidth}{!}{%
    \begin{tabular}{c|cccccc}
        \rowcolor{mygray}
        \thickhline
        Method & PIFuHD & PaMIR & ECON & GTA & SiTH & Ours \\
        \thickhline
        \multicolumn{1}{l|}{Geometry Quality} & 1.55 & 1.96 & \cellcolor{myyellow} 3.72 & 2.11 & 2.72 & \cellcolor{myorange} 4.71 \\
        \multicolumn{1}{l|}{Appearance Quality} & - & 1.42 & - & 2.65 & \cellcolor{myyellow}2.82 & \cellcolor{myorange} 4.59 \\
        \multicolumn{1}{l|}{Geometry Consistency} & 1.69 & 1.76 & 2.48 & 2.33 & \cellcolor{myyellow}2.79 & \cellcolor{myorange} 4.61 \\
        \multicolumn{1}{l|}{Appearance Consistency} & - & 1.77 & - & 2.16 & \cellcolor{myyellow}2.73 & \cellcolor{myorange} 4.68 \\
        \thickhline
    \end{tabular}}
\end{table}

\begin{figure*}
    \centering
    \includegraphics[width=\linewidth]{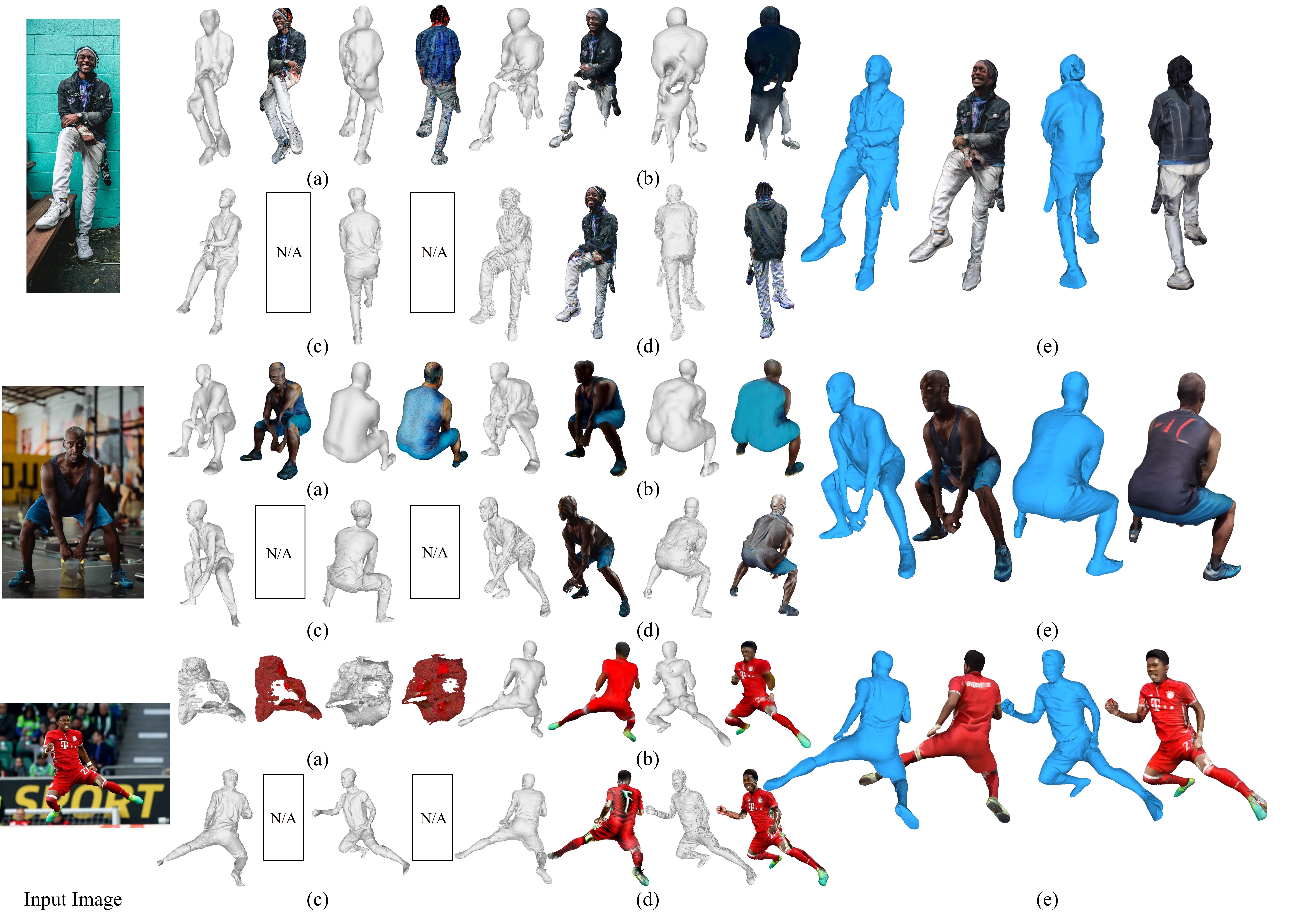}
    \caption{Qualitative comparison with optimization-based methods. We demonstrate the results of \textbf{(a)}~Magic123, \textbf{(b)}~Dreamgaussian, \textbf{(c)}~Chupa, \textbf{(d)}~TeCH and \textbf{(e)}~Ours.}
    \label{fig:comp_optim}
\end{figure*}

\noindent\textbf{User study} Given the limitations of quantitative metrics in assessing the realism and consistency of side and back views reconstructed from single-view input, we conducted a comprehensive user study to evaluate the geometry and appearance quality of five SOTA methods. Specifically, we collect 20 in-the-wild samples and 20 cases from SHHQ fashion dataset for evaluation. Following HumanNorm~\cite{huang2024humannorm}, we invite 20 volunteers to evaluate the color and normal video rendered from the reconstructed 3D humans. Participants were instructed to score each model on a 5-point scale (1 being the worst and 5 being the best) across four key dimensions:
\begin{itemize}
    \item To what extent does the human model exhibit the best geometry quality?
    \item To what extent does the human model exhibit the best appearance quality?
    \item To what extent does the novel view's geometry of the human body align with the reference image?
    \item To what extent does the novel view's appearance of the human body align with the reference image?
\end{itemize}

For methods that do not produce texture (PIFuHD and ECON), we only compare the geometry quality and consistency. The results in Tab.~\ref{tab:user_study} indicate that our method represents a significant advancement against SOTA methods, offering superior performance in both geometry and appearance reconstruction, as well as consistency across novel viewpoints.

\noindent\textbf{Comparison with optimization-based methods.}
To assess the efficacy of our approach relative to optimization-based methods, we conducted a comparative analysis of \methodname against several SDS-based techniques, Magic123, Dreamgaussian, Chupa, and TeCH. Following SiTH, we adopt the pose and text prompt generated by~\citep{li2022blip} as condition inputs due to the lack of direct image input support in Chupa. As illustrated in Fig.~\ref{fig:comp_optim}, Magic123 and Dreamgaussian exhibit significant limitations, primarily manifesting as incomplete human body reconstructions and implausible free-view textures. The reliance on text descriptions for conditioning proves insufficient for fine-grained control, resulting in geometries that deviate substantially from the reference inputs. TeCH, a method specifically designed for human reconstruction from a single image, while capable of producing complete human shapes, struggles with severe noise in geometric details and over-saturated textures. These artifacts are characteristic challenges inherent to SDS-based methodologies. In contrast, \methodname demonstrates superior performance by directly fusing multi-view 2D images in 3D space, enabling the preservation of geometry details at the pixel level while circumventing unrealistic texture. Note that TeCH requires $\sim$6 hours for optimization, \methodname generates high-quality textured meshes within merely 1 minute. We refer readers to Fig.~\ref{fig:shhq} and Fig.~\ref{fig:more_real} for more results generated by \methodname. 


\begin{figure}
    \centering
    \includegraphics[width=\linewidth]{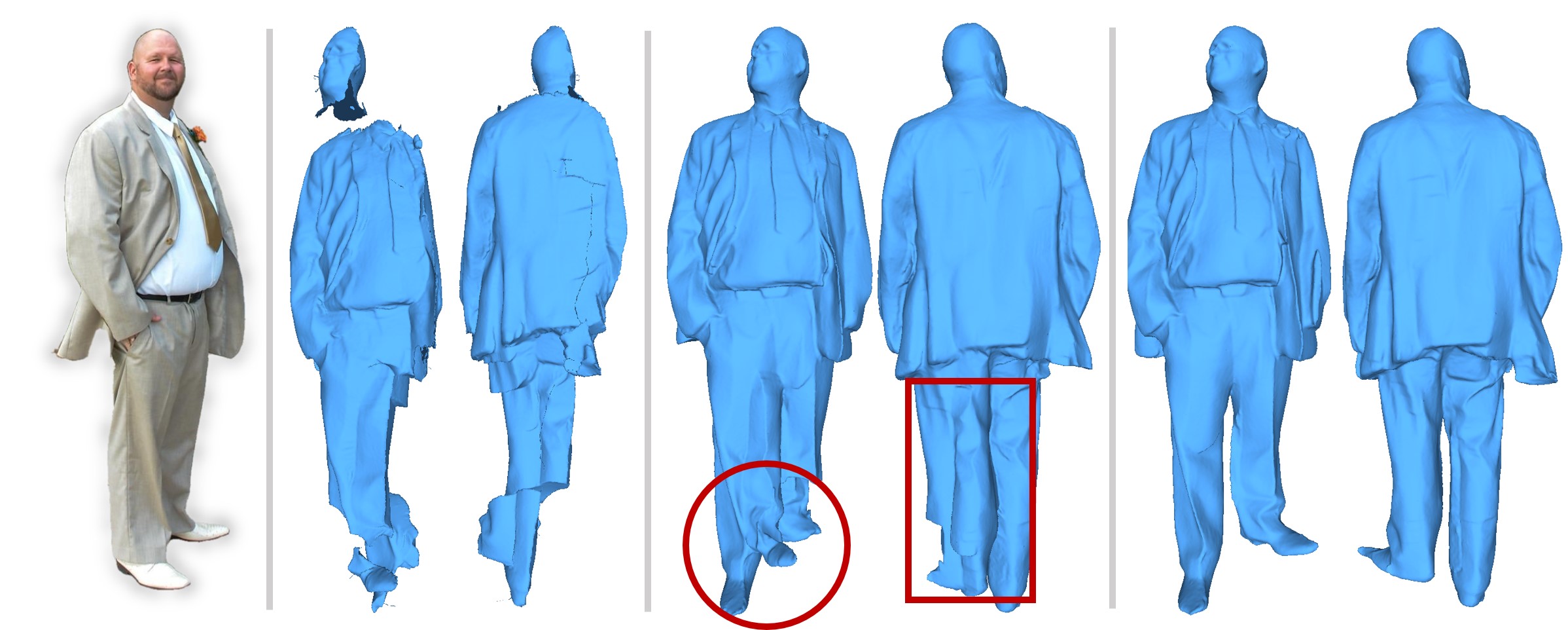}
\footnotesize\leftline{~~\quad~Input~\quad\qquad~view=2~\qquad\qquad~~view=4~\qquad\qquad~~view=6~(Ours)}
    \caption{Ablation of view number. Since normal maps lack depth information, optimizing geometry by only two or four views leads to an incomplete or unnatural human structure.}
    \label{fig:abla_view}
\end{figure}



\begin{figure}[t]
    \centering
     \vspace{-15pt}
    \includegraphics[width=\linewidth]{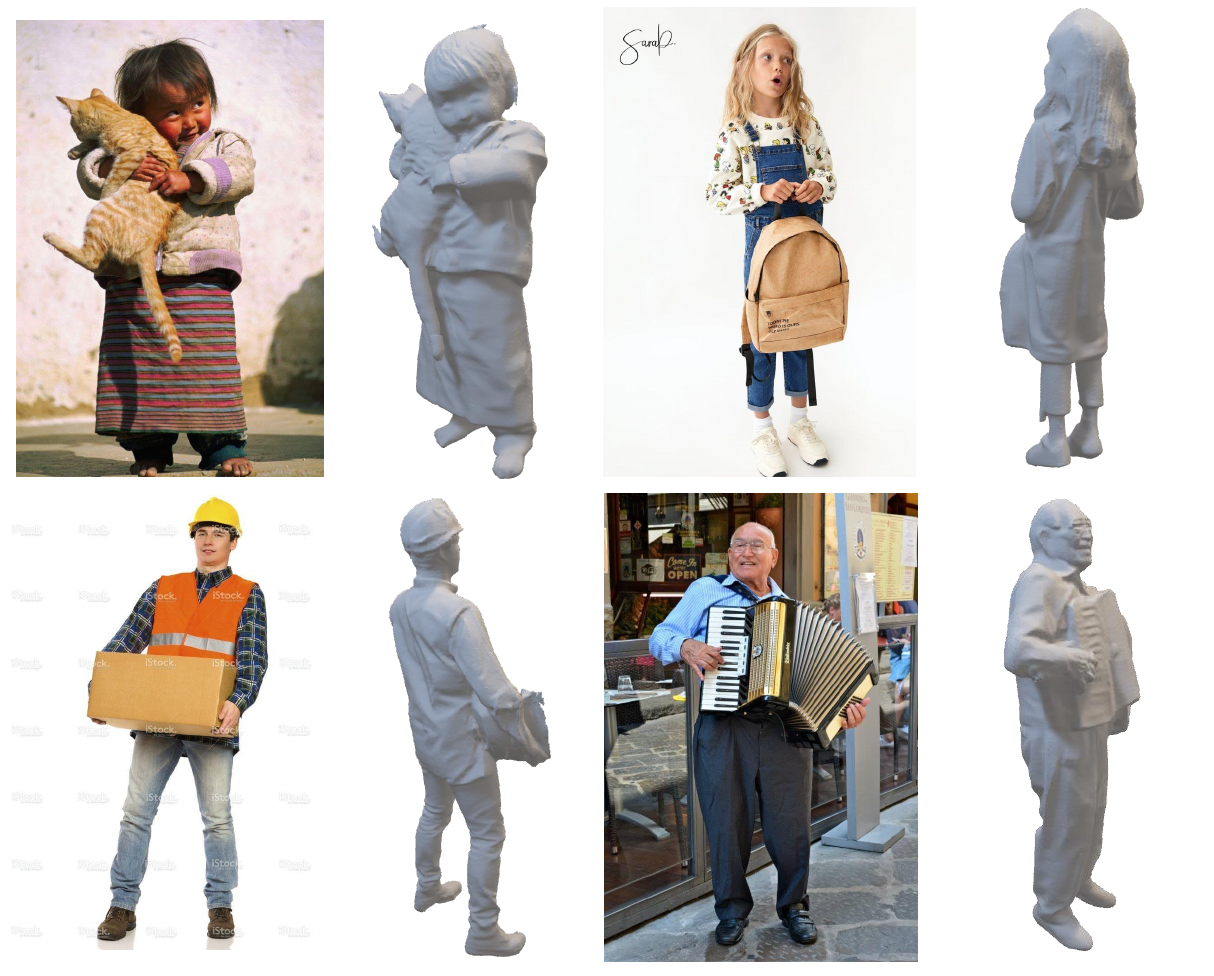}
    \small
    \vspace{-15pt}
    \caption{Reconstruction quality on object-occluded images.}
    \label{fig:object_occlusion}
    \vspace{-17pt}
\end{figure}

\noindent\textbf{Capability of handling occlusion.} We present the generated normal maps (back, left, and right views) and corresponding meshes of in-the-wild samples with various self-occlusion, as demonstrated in Fig.~\ref{fig:occlusion}.
To further illustrate the robustness of our approach, we also include examples of object-occluded scenarios in Fig. \ref{fig:object_occlusion}. The results show that our diffusion model can infer the correct human structure under both self-occlusions and object occlusions, enabling the reconstruction of high-quality 3D meshes even under such challenging conditions.

\begin{figure}[t]
    \centering
    \vspace{-5pt}
    \includegraphics[width=\linewidth]{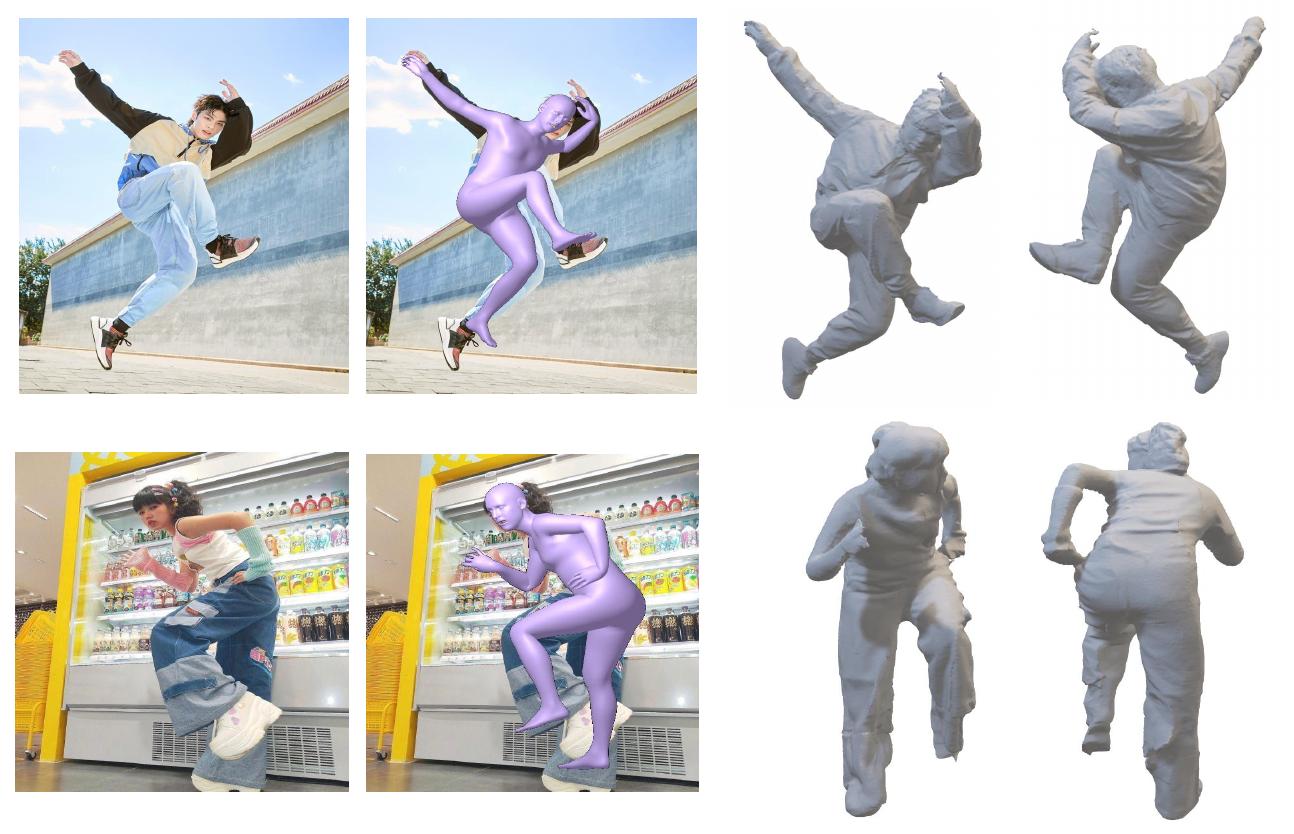}
    \vspace{-23pt}
    \caption{Robustness to SMPL-X estimation errors.}
    \label{fig:smplx_error}
\end{figure}

\begin{figure}[t]
    \centering
    \vspace{-5pt}
    \includegraphics[width=\linewidth]{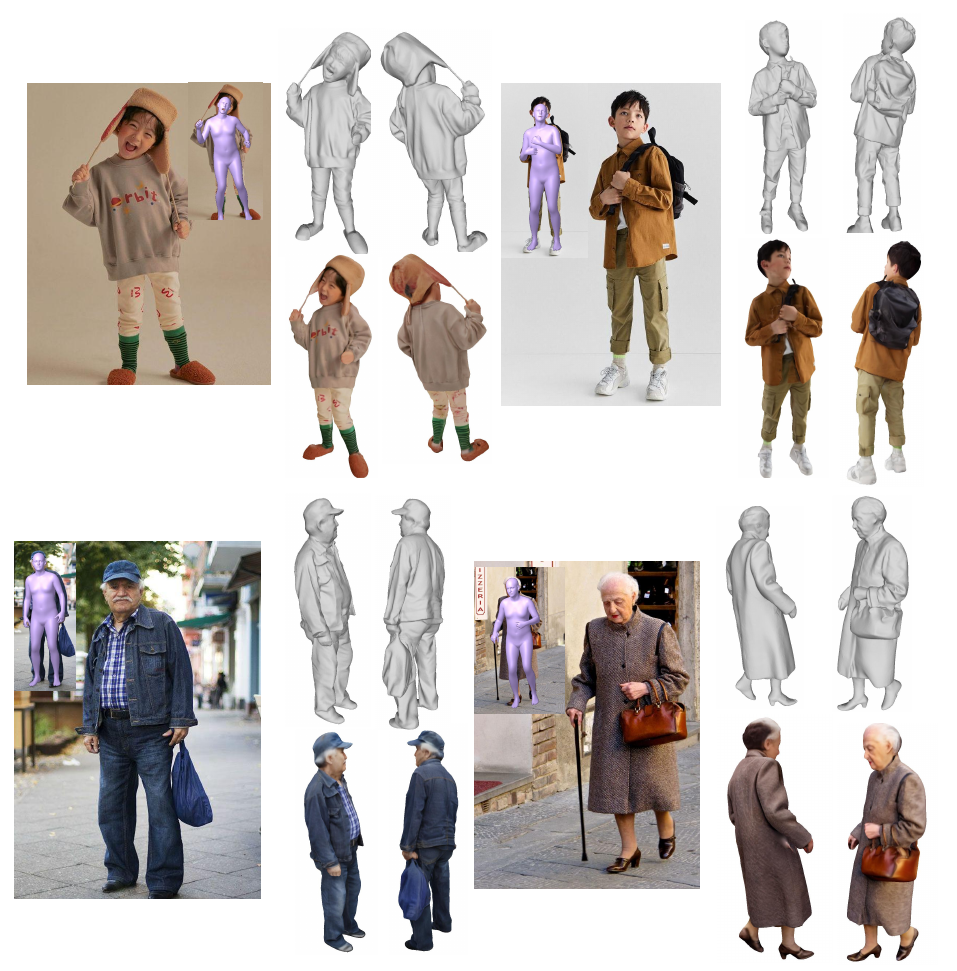}
    \vspace{-23pt}
    \caption{Performance with out-of-distribution pose estimation, like children and the elder.}
    \label{fig:child_elder}
\end{figure}

\noindent\textbf{Robustness to SMPL-X estimation.} The SMPL-X serves as a coarse anatomy guide, only required to be reasonably overlayed with the human body. Thus, our method could handle estimation error (Fig. \ref{fig:smplx_error}) to some extent and generalize to children or the elder in Fig. \ref{fig:child_elder}.

\begin{figure}[t]
    \centering
    \vspace{-5pt}
    \includegraphics[width=\linewidth]{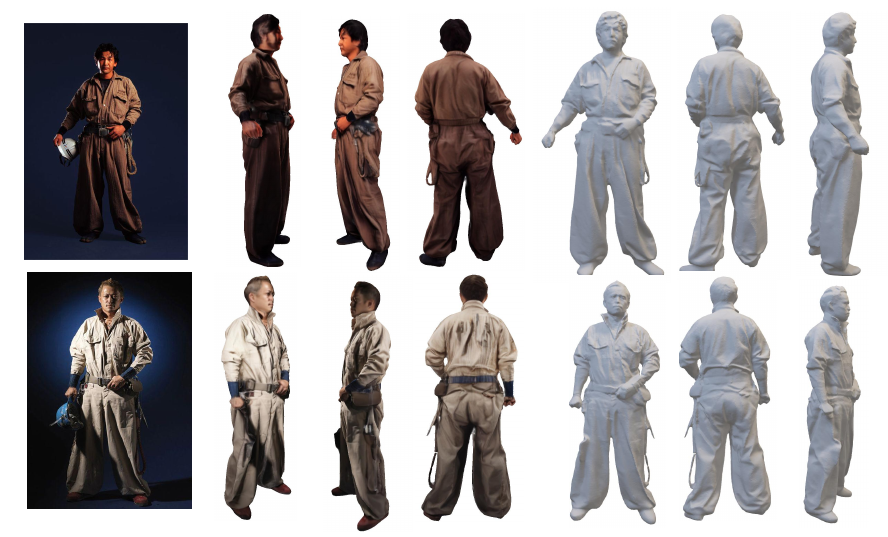}
    \vspace{-23pt}
    \caption{Robustness to shading and strong light.}
    \label{fig:lighting}
\end{figure}

\noindent\textbf{Robustness to lighting.} By incorporating varying lighting conditions using HDR maps from Poly Haven during training, our model demonstrates robustness to lighting variations, as illustrated in Fig. \ref{fig:lighting}.

\begin{figure}[t]
    \centering
    \vspace{-5pt}
    \includegraphics[width=\linewidth]{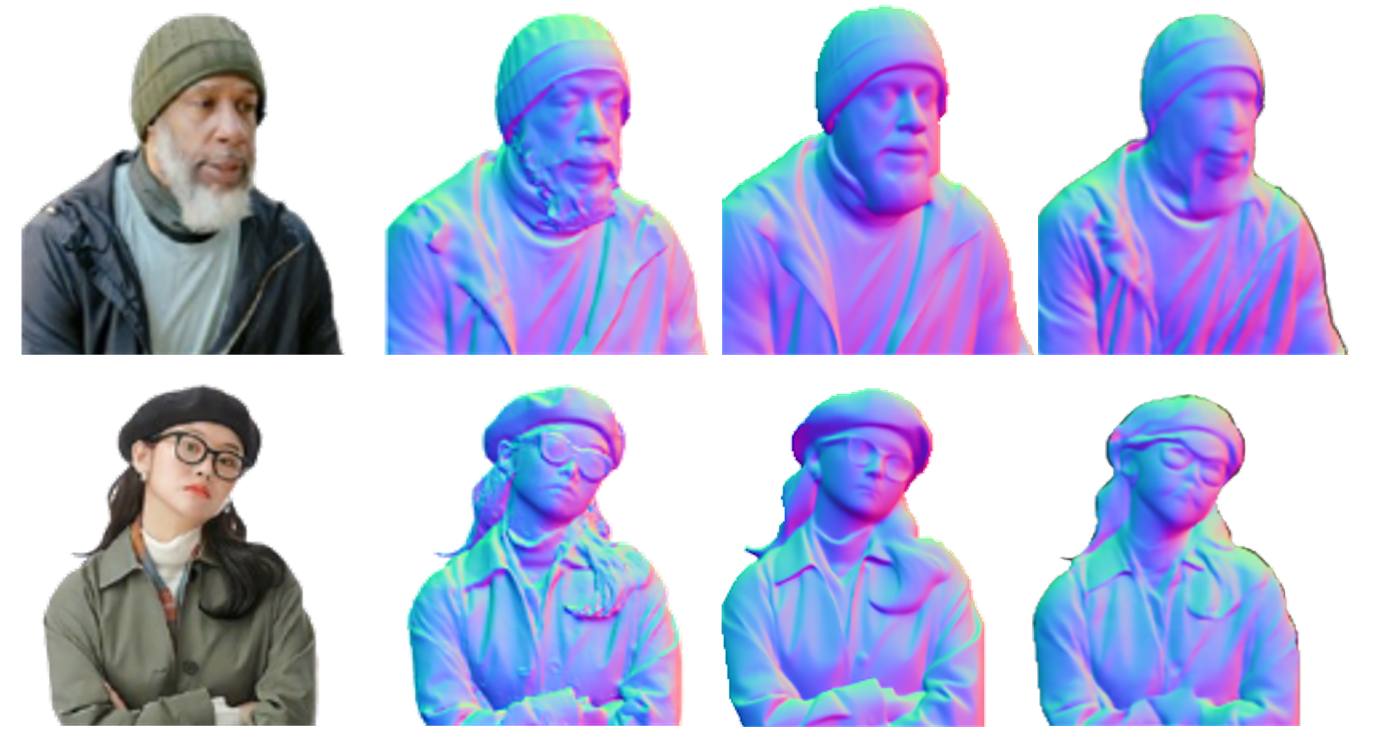}
    \vspace{-5pt}
    \small\leftline{~\qquad~~Input~~~~~\quad\qquad~Ours~\qquad~~Sapiens-2B~~~~\quad~~ECON}
    \caption{Comparisons of face normal estimation.}
    \label{fig:face_normal}
\end{figure}

\begin{figure*}
    \centering
    \includegraphics[width=0.7\linewidth]{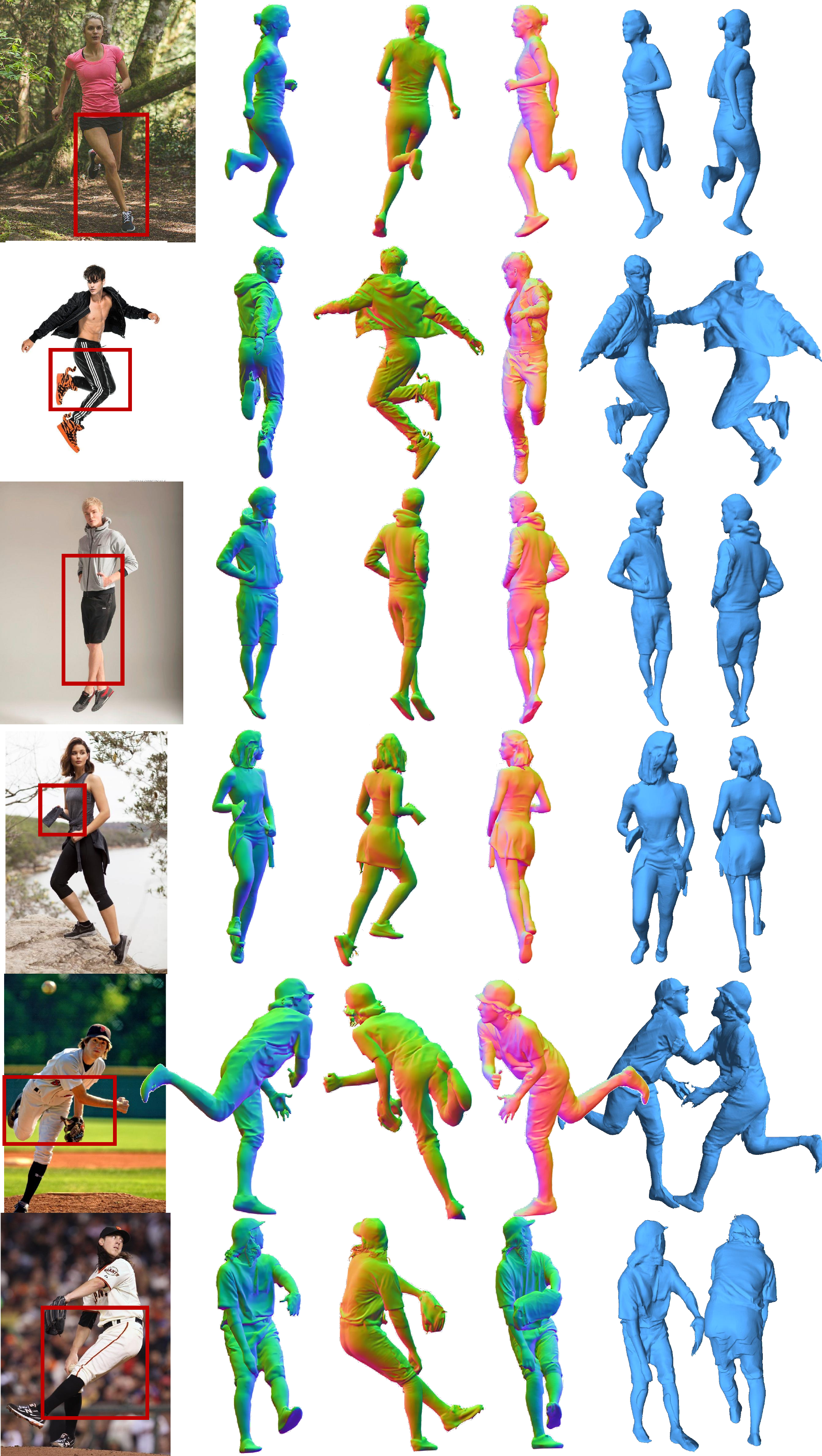}
    \caption{Reconstruction quality on \textcolor{red}{\textbf{self-occluded}} images. We present the generated back, left, and right views of normal maps and corresponding meshes. }
\label{fig:occlusion}
\end{figure*}

\noindent\textbf{Comparisons of face normal estimation.} As shown in Fig.~\ref{fig:face_normal}, our local face diffusion model generates facial normal images with significantly enhanced fine-grained details compared to ECON \cite{xiu2023econ} and SAPEIN-2B \cite{khirodkar2024sapiens}.

\begin{figure*}
    \centering
    \includegraphics[width=0.7\linewidth]{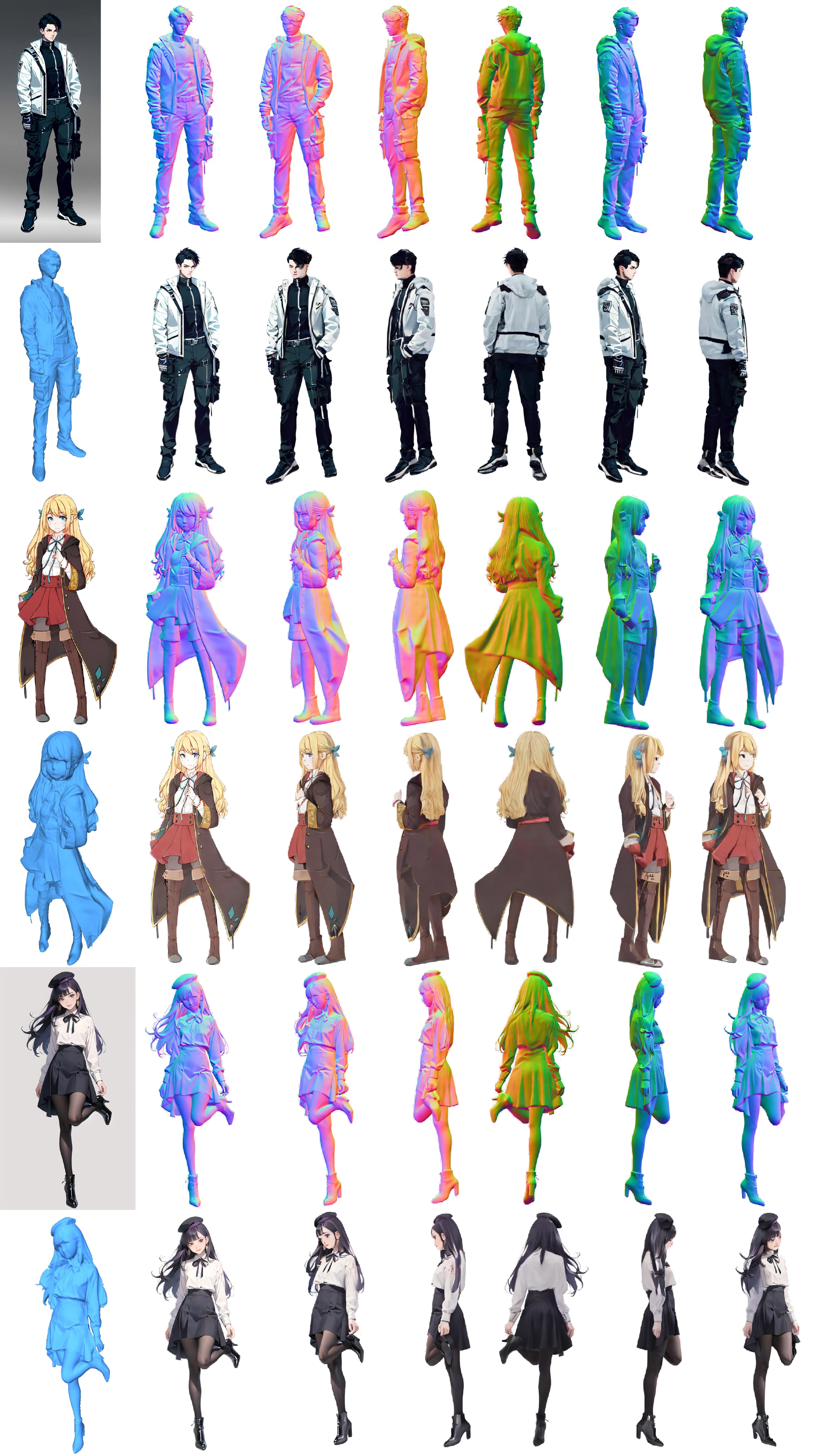}
    \caption{Generalization on anime characters. We present the generated multiview color and normal images and corresponding meshes (in blue). }
\label{fig:cartoon}
\end{figure*}

\begin{figure*}
    \centering
    \includegraphics[width=0.8\linewidth]{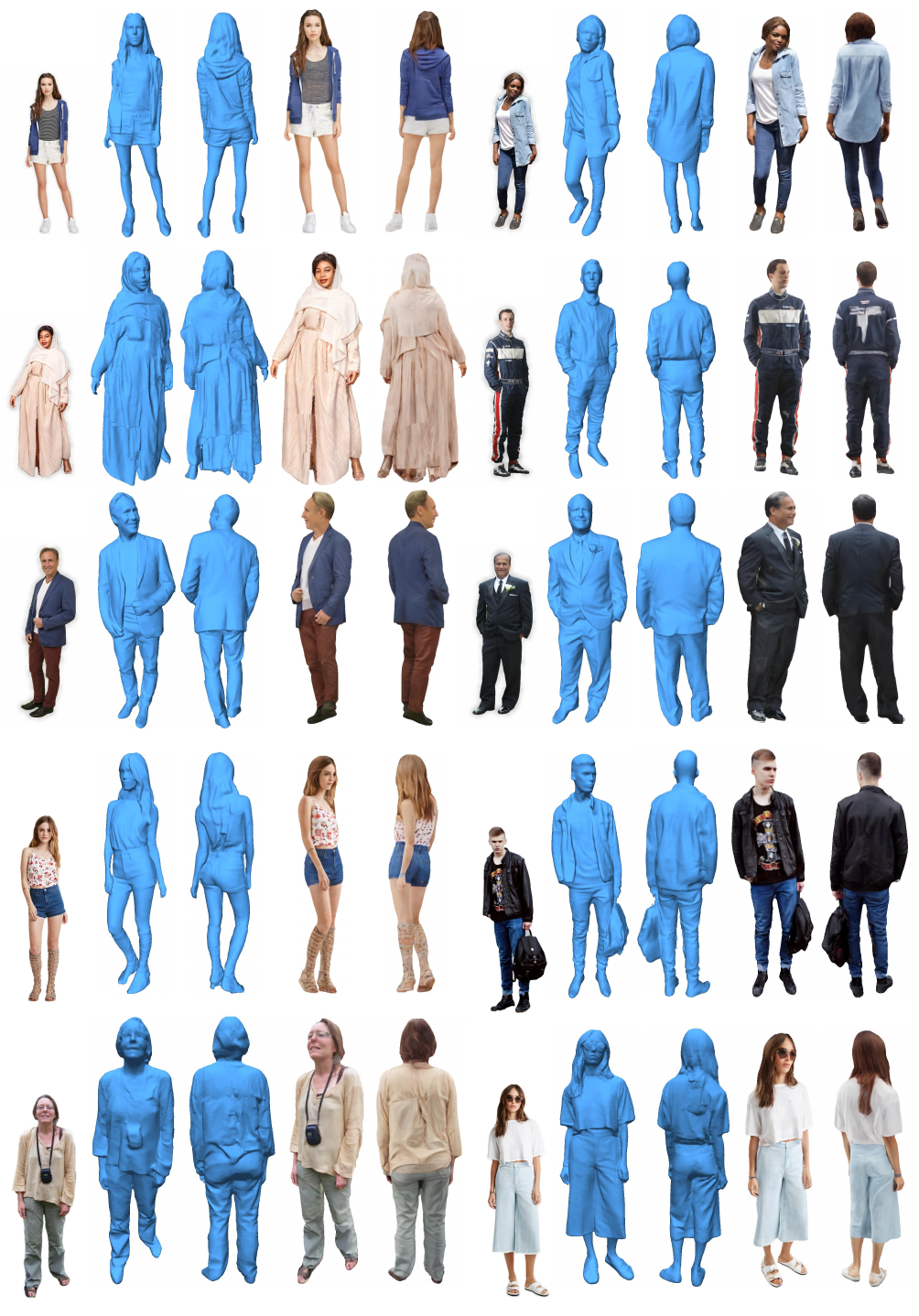}
    \caption{More results on SHHQ dataset.}
\label{fig:shhq}
\end{figure*}

\begin{figure*}
    \centering
    \includegraphics[width=\linewidth]{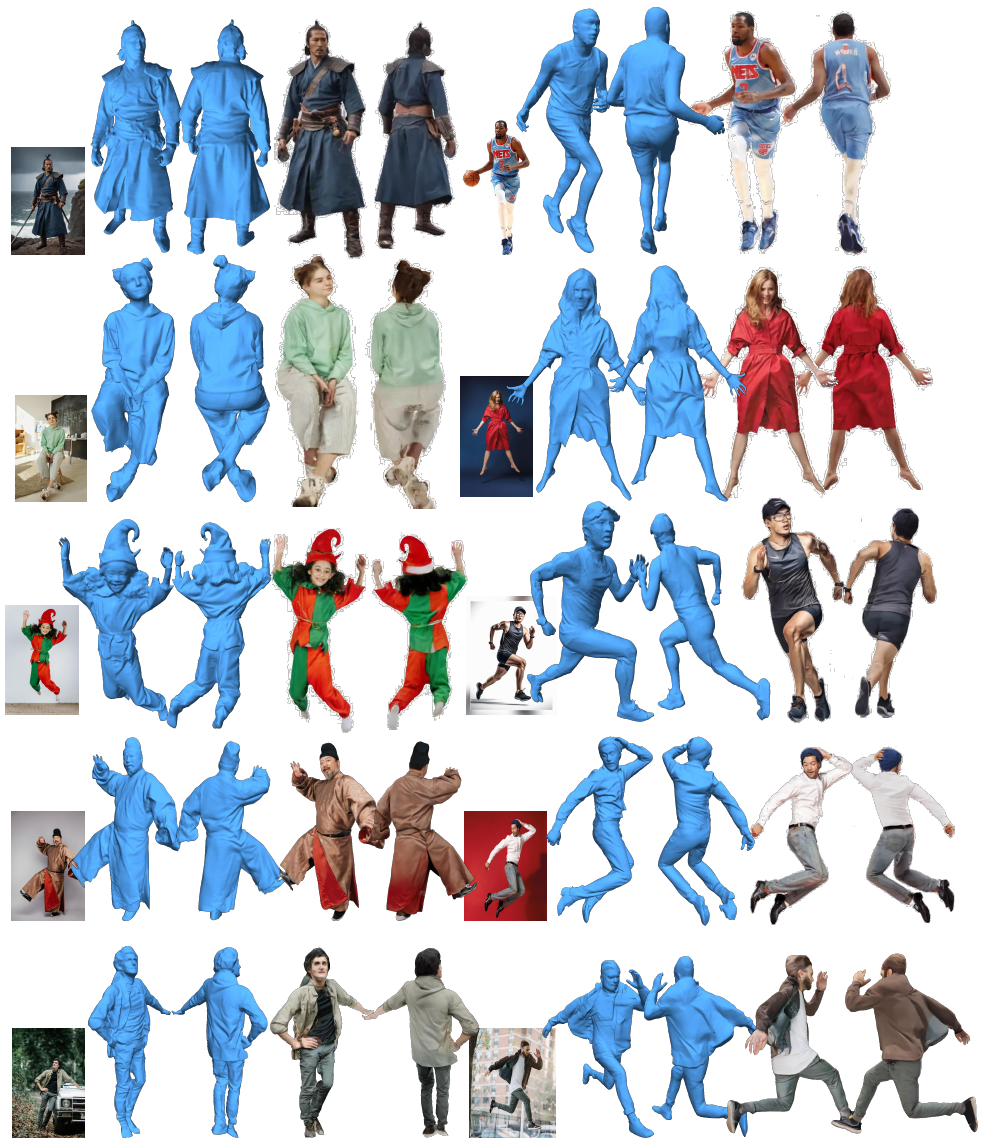}
    \caption{More results on in-the-wild data.}
\label{fig:more_real}
\end{figure*}

\noindent\textbf{Generalization on anime characters.} Our model, trained with only realistic human scans, exhibits excellent generalization on anime or hand-drawn style character images, as shown in Fig.~\ref{fig:cartoon}. This is because our method is adapted from the Stable Diffusion~\cite{stable_diffusion} model, which has been trained on images of various styles. Thus, our method maintains the ability to generalize images of different domains.


\section{Ethics statement}
While \methodname aims to provide users with an advanced tool for single-image full-body 3D human model reconstruction, we acknowledge the potential for misuse, particularly in creating deceptive content. This ethical concern extends beyond our specific method to the broader field of generative modeling. As researchers and developers in 3D reconstruction and generative AI, we have a responsibility to continually address these ethical implications. We encourage ongoing dialogue and the development of safeguards to mitigate potential harm while advancing the technology responsibly. Users of \methodname and similar tools should be aware of these ethical considerations and use the technology in accordance with applicable laws and ethical guidelines.

%% file: main.bib
@String(IJCV = {Int. J. Comput. Vis.})

@String(CVPR= {IEEE Conf. Comput. Vis. Pattern Recog.})

@String(ICCV= {Int. Conf. Comput. Vis.})

@String(ECCV= {Eur. Conf. Comput. Vis.})

@String(TOG= {ACM Trans. Graph.})

@String(IJCV  = {IJCV})

@String(CVPR  = {CVPR})

@String(ICCV  = {ICCV})

@String(ECCV  = {ECCV})

@String(TOG   = {ACM TOG})

@inproceedings{rombach2022high,
	title        = {High-resolution image synthesis with latent diffusion models},
	author       = {Rombach, Robin and Blattmann, Andreas and Lorenz, Dominik and Esser, Patrick and Ommer, Bj{\"o}rn},
	year         = 2022,
	booktitle    = {CVPR},
	pages        = {10684--10695}
}

@article{poole2022dreamfusion,
	title        = {Dreamfusion: Text-to-3d using 2d diffusion},
	author       = {Poole, Ben and Jain, Ajay and Barron, Jonathan T and Mildenhall, Ben},
	year         = 2022,
	journal      = {arXiv preprint arXiv:2209.14988}
}

@article{liu2023syncdreamer,
	title        = {SyncDreamer: Generating Multiview-consistent Images from a Single-view Image},
	author       = {Liu, Yuan and Lin, Cheng and Zeng, Zijiao and Long, Xiaoxiao and Liu, Lingjie and Komura, Taku and Wang, Wenping},
	year         = 2023,
	journal      = {arXiv preprint arXiv:2309.03453}
}

@article{tang2024mvdiffusionpp,
  title={MVDiffusion++: A Dense High-resolution Multi-view Diffusion Model for Single or Sparse-view 3D Object Reconstruction},
  author={Tang, Shitao and Chen, Jiacheng and Wang, Dilin and Tang, Chengzhou and Zhang, Fuyang and Fan, Yuchen and Chandra, Vikas and Furukawa, Yasutaka and Ranjan, Rakesh},
  journal={arXiv preprint arXiv:2402.12712},
  year={2024}
}

@article{voleti2024sv3d,
  title={Sv3d: Novel multi-view synthesis and 3d generation from a single image using latent video diffusion},
  author={Voleti, Vikram and Yao, Chun-Han and Boss, Mark and Letts, Adam and Pankratz, David and Tochilkin, Dmitry and Laforte, Christian and Rombach, Robin and Jampani, Varun},
  journal={arXiv preprint arXiv:2403.12008},
  year={2024}
}

@article{liao2023tada,
	title        = {TADA! Text to Animatable Digital Avatars},
	author       = {Liao, Tingting and Yi, Hongwei and Xiu, Yuliang and Tang, Jiaxaing and Huang, Yangyi and Thies, Justus and Black, Michael J},
	year         = 2023,
	journal      = {arXiv preprint arXiv:2308.10899}
}

@inproceedings{yu2021pixelnerf,
	title        = {pixelnerf: Neural radiance fields from one or few images},
	author       = {Yu, Alex and Ye, Vickie and Tancik, Matthew and Kanazawa, Angjoo},
	year         = 2021,
	booktitle    = {CVPR},
	pages        = {4578--4587}
}

@inproceedings{li2022blip,
	title        = {Blip: Bootstrapping language-image pre-training for unified vision-language understanding and generation},
	author       = {Li, Junnan and Li, Dongxu and Xiong, Caiming and Hoi, Steven},
	year         = 2022,
	booktitle    = {ICML},
	pages        = {12888--12900},
	organization = {PMLR}
}

@InProceedings{stable_diffusion,
    author    = {Rombach, Robin and Blattmann, Andreas and Lorenz, Dominik and Esser, Patrick and Ommer, Bj\"orn},
    title     = {High-Resolution Image Synthesis With Latent Diffusion Models},
    booktitle = {Proceedings of the IEEE/CVF Conference on Computer Vision and Pattern Recognition (CVPR)},
    month     = {June},
    year      = {2022},
    pages     = {10684-10695}
}

@article{li2024era3d,
  title={Era3D: High-Resolution Multiview Diffusion using Efficient Row-wise Attention},
  author={Li, Peng and Liu, Yuan and Long, Xiaoxiao and Zhang, Feihu and Lin, Cheng and Li, Mengfei and Qi, Xingqun and Zhang, Shanghang and Luo, Wenhan and Tan, Ping and others},
  journal={arXiv preprint arXiv:2405.11616},
  year={2024}
}

@InProceedings{wonder3d,
    author    = {Long, Xiaoxiao and Guo, Yuan-Chen and Lin, Cheng and Liu, Yuan and Dou, Zhiyang and Liu, Lingjie and Ma, Yuexin and Zhang, Song-Hai and Habermann, Marc and Theobalt, Christian and Wang, Wenping},
    title     = {Wonder3D: Single Image to 3D using Cross-Domain Diffusion},
    booktitle = {Proceedings of the IEEE/CVF Conference on Computer Vision and Pattern Recognition (CVPR)},
    month     = {June},
    year      = {2024},
    pages     = {9970-9980}
}

@inproceedings{chibane2020ifnet,
  title={Implicit functions in feature space for 3d shape reconstruction and completion},
  author={Chibane, Julian and Alldieck, Thiemo and Pons-Moll, Gerard},
  booktitle={Proceedings of the IEEE/CVF conference on computer vision and pattern recognition},
  pages={6970--6981},
  year={2020}
}

@article{gropp2020implicit,
  title={Implicit geometric regularization for learning shapes},
  author={Gropp, Amos and Yariv, Lior and Haim, Niv and Atzmon, Matan and Lipman, Yaron},
  journal={arXiv preprint arXiv:2002.10099},
  year={2020}
}

@inproceedings{ma2021pixel,
  title={Pixel codec avatars},
  author={Ma, Shugao and Simon, Tomas and Saragih, Jason and Wang, Dawei and Li, Yuecheng and De La Torre, Fernando and Sheikh, Yaser},
  booktitle={Proceedings of the IEEE/CVF Conference on Computer Vision and Pattern Recognition},
  pages={64--73},
  year={2021}
}

@inproceedings{orts2016holoportation,
  title={Holoportation: Virtual 3d teleportation in real-time},
  author={Orts-Escolano, Sergio and Rhemann, Christoph and Fanello, Sean and Chang, Wayne and Kowdle, Adarsh and Degtyarev, Yury and Kim, David and Davidson, Philip L and Khamis, Sameh and Dou, Mingsong and others},
  booktitle={Proceedings of the 29th annual symposium on user interface software and technology},
  pages={741--754},
  year={2016}
}

@article{poisson_recon,
  title={Screened poisson surface reconstruction},
  author={Kazhdan, Michael and Hoppe, Hugues},
  journal={ACM Transactions on Graphics (ToG)},
  volume={32},
  number={3},
  pages={1--13},
  year={2013},
  publisher={ACM New York, NY, USA}
}

@InProceedings{occupancy_network,
author = {Mescheder, Lars and Oechsle, Michael and Niemeyer, Michael and Nowozin, Sebastian and Geiger, Andreas},
title = {Occupancy Networks: Learning 3D Reconstruction in Function Space},
booktitle = {Proceedings of the IEEE/CVF Conference on Computer Vision and Pattern Recognition (CVPR)},
month = {June},
year = {2019}
}

@InProceedings{deepsdf,
author = {Park, Jeong Joon and Florence, Peter and Straub, Julian and Newcombe, Richard and Lovegrove, Steven},
title = {DeepSDF: Learning Continuous Signed Distance Functions for Shape Representation},
booktitle = {Proceedings of the IEEE/CVF Conference on Computer Vision and Pattern Recognition (CVPR)},
month = {June},
year = {2019}
}

@article{palfinger2022continuous,
  title={Continuous remeshing for inverse rendering},
  author={Palfinger, Werner},
  journal={Computer Animation and Virtual Worlds},
  volume={33},
  number={5},
  pages={e2101},
  year={2022},
  publisher={Wiley Online Library}
}

@InProceedings{D-IF,
    author    = {Yang, Xueting and Luo, Yihao and Xiu, Yuliang and Wang, Wei and Xu, Hao and Fan, Zhaoxin},
    title     = {D-IF: Uncertainty-aware Human Digitization via Implicit Distribution Field},
    booktitle = {Proceedings of the IEEE/CVF International Conference on Computer Vision (ICCV)},
    month     = {October},
    year      = {2023},
    pages     = {9122-9132}
}

@inproceedings{saito2019pifu,
  title={Pifu: Pixel-aligned implicit function for high-resolution clothed human digitization},
  author={Saito, Shunsuke and Huang, Zeng and Natsume, Ryota and Morishima, Shigeo and Kanazawa, Angjoo and Li, Hao},
  booktitle={Proceedings of the IEEE/CVF international conference on computer vision},
  pages={2304--2314},
  year={2019}
}

@inproceedings{saito2020pifuhd,
  title={Pifuhd: Multi-level pixel-aligned implicit function for high-resolution 3d human digitization},
  author={Saito, Shunsuke and Simon, Tomas and Saragih, Jason and Joo, Hanbyul},
  booktitle={Proceedings of the IEEE/CVF conference on computer vision and pattern recognition},
  pages={84--93},
  year={2020}
}

@article{zheng2021pamir,
  title={Pamir: Parametric model-conditioned implicit representation for image-based human reconstruction},
  author={Zheng, Zerong and Yu, Tao and Liu, Yebin and Dai, Qionghai},
  journal={IEEE transactions on pattern analysis and machine intelligence},
  volume={44},
  number={6},
  pages={3170--3184},
  year={2021},
  publisher={IEEE}
}

@InProceedings{arch,
author = {Huang, Zeng and Xu, Yuanlu and Lassner, Christoph and Li, Hao and Tung, Tony},
title = {ARCH: Animatable Reconstruction of Clothed Humans},
booktitle = {Proceedings of the IEEE/CVF Conference on Computer Vision and Pattern Recognition (CVPR)},
month = {June},
year = {2020}
}

@InProceedings{arch++,
    author    = {He, Tong and Xu, Yuanlu and Saito, Shunsuke and Soatto, Stefano and Tung, Tony},
    title     = {ARCH++: Animation-Ready Clothed Human Reconstruction Revisited},
    booktitle = {Proceedings of the IEEE/CVF International Conference on Computer Vision (ICCV)},
    month     = {October},
    year      = {2021},
    pages     = {11046-11056}
}

@InProceedings{function4d,
    author    = {Yu, Tao and Zheng, Zerong and Guo, Kaiwen and Liu, Pengpeng and Dai, Qionghai and Liu, Yebin},
    title     = {Function4D: Real-Time Human Volumetric Capture From Very Sparse Consumer RGBD Sensors},
    booktitle = {Proceedings of the IEEE/CVF Conference on Computer Vision and Pattern Recognition (CVPR)},
    month     = {June},
    year      = {2021},
    pages     = {5746-5756}
}

@inproceedings{xiu2022icon,
  title={Icon: Implicit clothed humans obtained from normals},
  author={Xiu, Yuliang and Yang, Jinlong and Tzionas, Dimitrios and Black, Michael J},
  booktitle={2022 IEEE/CVF Conference on Computer Vision and Pattern Recognition (CVPR)},
  pages={13286--13296},
  year={2022},
  organization={IEEE}
}

@inproceedings{xiu2023econ,
  title={Econ: Explicit clothed humans optimized via normal integration},
  author={Xiu, Yuliang and Yang, Jinlong and Cao, Xu and Tzionas, Dimitrios and Black, Michael J},
  booktitle={Proceedings of the IEEE/CVF conference on computer vision and pattern recognition},
  pages={512--523},
  year={2023}
}

@InProceedings{phorhuman,
    author    = {Alldieck, Thiemo and Zanfir, Mihai and Sminchisescu, Cristian},
    title     = {Photorealistic Monocular 3D Reconstruction of Humans Wearing Clothing},
    booktitle = {Proceedings of the IEEE/CVF Conference on Computer Vision and Pattern Recognition (CVPR)},
    month     = {June},
    year      = {2022},
    pages     = {1506-1515}
}

@article{gta,
  title={Global-correlated 3d-decoupling transformer for clothed avatar reconstruction},
  author={Zhang, Zechuan and Sun, Li and Yang, Zongxin and Chen, Ling and Yang, Yi},
  journal={Advances in Neural Information Processing Systems},
  volume={36},
  year={2024}
}

@inproceedings{zhang2024sifu,
  title={Sifu: Side-view conditioned implicit function for real-world usable clothed human reconstruction},
  author={Zhang, Zechuan and Yang, Zongxin and Yang, Yi},
  booktitle={Proceedings of the IEEE/CVF Conference on Computer Vision and Pattern Recognition},
  pages={9936--9947},
  year={2024}
}

@inproceedings{ho2024sith,
  title={Sith: Single-view textured human reconstruction with image-conditioned diffusion},
  author={Ho, I and Song, Jie and Hilliges, Otmar and others},
  booktitle={Proceedings of the IEEE/CVF Conference on Computer Vision and Pattern Recognition},
  pages={538--549},
  year={2024}
}

@inproceedings{huang2024tech,
  title={Tech: Text-guided reconstruction of lifelike clothed humans},
  author={Huang, Yangyi and Yi, Hongwei and Xiu, Yuliang and Liao, Tingting and Tang, Jiaxiang and Cai, Deng and Thies, Justus},
  booktitle={2024 International Conference on 3D Vision (3DV)},
  pages={1531--1542},
  year={2024},
  organization={IEEE}
}

@InProceedings{chupa,
    author    = {Kim, Byungjun and Kwon, Patrick and Lee, Kwangho and Lee, Myunggi and Han, Sookwan and Kim, Daesik and Joo, Hanbyul},
    title     = {Chupa: Carving 3D Clothed Humans from Skinned Shape Priors using 2D Diffusion Probabilistic Models},
    booktitle = {Proceedings of the IEEE/CVF International Conference on Computer Vision (ICCV)},
    month     = {October},
    year      = {2023},
    pages     = {15965-15976}
}

@inproceedings{huang2024humannorm,
  title={Humannorm: Learning normal diffusion model for high-quality and realistic 3d human generation},
  author={Huang, Xin and Shao, Ruizhi and Zhang, Qi and Zhang, Hongwen and Feng, Ying and Liu, Yebin and Wang, Qing},
  booktitle={Proceedings of the IEEE/CVF Conference on Computer Vision and Pattern Recognition},
  pages={4568--4577},
  year={2024}
}

@inproceedings{yang2024hilo,
  title={HiLo: Detailed and Robust 3D Clothed Human Reconstruction with High-and Low-Frequency Information of Parametric Models},
  author={Yang, Yifan and Liu, Dong and Zhang, Shuhai and Deng, Zeshuai and Huang, Zixiong and Tan, Mingkui},
  booktitle={Proceedings of the IEEE/CVF Conference on Computer Vision and Pattern Recognition},
  pages={10671--10681},
  year={2024}
}

@InProceedings{bodynet,
author = {Varol, Gul and Ceylan, Duygu and Russell, Bryan and Yang, Jimei and Yumer, Ersin and Laptev, Ivan and Schmid, Cordelia},
title = {BodyNet: Volumetric Inference of 3D Human Body Shapes},
booktitle = {Proceedings of the European Conference on Computer Vision (ECCV)},
month = {September},
year = {2018}
}

@InProceedings{deephuman,
author = {Zheng, Zerong and Yu, Tao and Wei, Yixuan and Dai, Qionghai and Liu, Yebin},
title = {DeepHuman: 3D Human Reconstruction From a Single Image},
booktitle = {Proceedings of the IEEE/CVF International Conference on Computer Vision (ICCV)},
month = {October},
year = {2019}
}

@InProceedings{SiCloPe,
author = {Natsume, Ryota and Saito, Shunsuke and Huang, Zeng and Chen, Weikai and Ma, Chongyang and Li, Hao and Morishima, Shigeo},
title = {SiCloPe: Silhouette-Based Clothed People},
booktitle = {Proceedings of the IEEE/CVF Conference on Computer Vision and Pattern Recognition (CVPR)},
month = {June},
year = {2019}
}

@InProceedings{Moulding_Humans,
author = {Gabeur, Valentin and Franco, Jean-Sebastien and Martin, Xavier and Schmid, Cordelia and Rogez, Gregory},
title = {Moulding Humans: Non-Parametric 3D Human Shape Estimation From Single Images},
booktitle = {Proceedings of the IEEE/CVF International Conference on Computer Vision (ICCV)},
month = {October},
year = {2019}
}

@InProceedings{FACSIMILE,
author = {Smith, David and Loper, Matthew and Hu, Xiaochen and Mavroidis, Paris and Romero, Javier},
title = {FACSIMILE: Fast and Accurate Scans From an Image in Less Than a Second},
booktitle = {Proceedings of the IEEE/CVF International Conference on Computer Vision (ICCV)},
month = {October},
year = {2019}
}

@article{he2020geopifu,
  title={Geo-pifu: Geometry and pixel aligned implicit functions for single-view human reconstruction},
  author={He, Tong and Collomosse, John and Jin, Hailin and Soatto, Stefano},
  journal={Advances in Neural Information Processing Systems},
  volume={33},
  pages={9276--9287},
  year={2020}
}

@InProceedings{visrecon,
    author    = {Zheng, Ruichen and Li, Peng and Wang, Haoqian and Yu, Tao},
    title     = {Learning Visibility Field for Detailed 3D Human Reconstruction and Relighting},
    booktitle = {Proceedings of the IEEE/CVF Conference on Computer Vision and Pattern Recognition (CVPR)},
    month     = {June},
    year      = {2023},
    pages     = {216-226}
}

@inproceedings{bini2022cao,
  title={Bilateral Normal Integration},
  author={Cao, Xu and Santo, Hiroaki and Shi, Boxin and Okura, Fumio and Matsushita, Yasuyuki},
  booktitle=ECCV,
  year={2022}
}

@InProceedings{2k2k,
    author    = {Han, Sang-Hun and Park, Min-Gyu and Yoon, Ju Hong and Kang, Ju-Mi and Park, Young-Jae and Jeon, Hae-Gon},
    title     = {High-Fidelity 3D Human Digitization From Single 2K Resolution Images},
    booktitle = {Proceedings of the IEEE/CVF Conference on Computer Vision and Pattern Recognition (CVPR)},
    month     = {June},
    year      = {2023},
    pages     = {12869-12879}
}

@InProceedings{Tex2Shape,
author = {Alldieck, Thiemo and Pons-Moll, Gerard and Theobalt, Christian and Magnor, Marcus},
title = {Tex2Shape: Detailed Full Human Body Geometry From a Single Image},
booktitle = {Proceedings of the IEEE/CVF International Conference on Computer Vision (ICCV)},
month = {October},
year = {2019}
}

@InProceedings{CustomHumans,
    author    = {Ho, Hsuan-I and Xue, Lixin and Song, Jie and Hilliges, Otmar},
    title     = {Learning Locally Editable Virtual Humans},
    booktitle = {Proceedings of the IEEE/CVF Conference on Computer Vision and Pattern Recognition (CVPR)},
    month     = {June},
    year      = {2023},
    pages     = {21024-21035}
}

@InProceedings{Ma_cape,
author = {Ma, Qianli and Yang, Jinlong and Ranjan, Anurag and Pujades, Sergi and Pons-Moll, Gerard and Tang, Siyu and Black, Michael J.},
title = {Learning to Dress 3D People in Generative Clothing},
booktitle = {Proceedings of the IEEE/CVF Conference on Computer Vision and Pattern Recognition (CVPR)},
month = {June},
year = {2020}
}

@article{deng2018menpo,
  title={The Menpo benchmark for multi-pose 2D and 3D facial landmark localisation and tracking},
  author={Deng, Jiankang and Roussos, Anastasios and Chrysos, Grigorios and Ververas, Evangelos and Kotsia, Irene and Shen, Jie and Zafeiriou, Stefanos},
  journal={IJCV},
  year={2018}
}

@inproceedings{zhangUnreasonableEffectivenessDeep2018,
  author       = {Richard Zhang and
                  Phillip Isola and
                  Alexei A. Efros and
                  Eli Shechtman and
                  Oliver Wang},
  title        = {The Unreasonable Effectiveness of Deep Features as a Perceptual Metric},
  booktitle    = {{CVPR}},
  pages        = {586--595},
  publisher    = {Computer Vision Foundation / {IEEE} Computer Society},
  year         = {2018}
}

@article{wang2004image,
  author       = {Zhou Wang and
                  Alan C. Bovik and
                  Hamid R. Sheikh and
                  Eero P. Simoncelli},
  title        = {Image quality assessment: from error visibility to structural similarity},
  journal      = {{IEEE} Trans. Image Process.},
  volume       = {13},
  number       = {4},
  pages        = {600--612},
  year         = {2004}
}

@inproceedings{zhang2018unreasonable,
  author       = {Richard Zhang and
                  Phillip Isola and
                  Alexei A. Efros and
                  Eli Shechtman and
                  Oliver Wang},
  title        = {The Unreasonable Effectiveness of Deep Features as a Perceptual Metric},
  booktitle    = {{CVPR}},
  pages        = {586--595},
  publisher    = {Computer Vision Foundation / {IEEE} Computer Society},
  year         = {2018}
}

@misc{rembg,
  author       = {Daniel Gatis},
  title        = {rembg},
  year         = {},
  url          = {https://github.com/danielgatis/rembg},
  note         = {https://github.com/danielgatis/rembg}
}

@inproceedings{khirodkar2024sapiens,
  title={Sapiens: Foundation for human vision models},
  author={Khirodkar, Rawal and Bagautdinov, Timur and Martinez, Julieta and Zhaoen, Su and James, Austin and Selednik, Peter and Anderson, Stuart and Saito, Shunsuke},
  booktitle={European Conference on Computer Vision},
  pages={206--228},
  year={2024},
  organization={Springer}
}
